\documentclass[conference]{IEEEtran}
\usepackage{amsfonts}

\usepackage{adjustbox}
\usepackage{algorithmicx}
\usepackage{algpseudocode}
\usepackage[ruled]{algorithm}
\usepackage{graphicx}
\usepackage{caption}
\usepackage{graphicx,times,amsmath} 
\usepackage{caption}
\usepackage{subcaption}
\usepackage[rightcaption]{sidecap}
\usepackage{caption}
\usepackage{multirow}
\usepackage{hhline}
\usepackage{amsmath}
\usepackage{epstopdf}
\usepackage{tabularx}
\usepackage{float}

\usepackage[font={small}]{caption}
\usepackage{cite}

\usepackage{fancyhdr}

\fancypagestyle{pageStyleOne}{%
\normalfont\footnotesize
    \fancyhf{}
\fancyhead[L]{\fontsize{8}{8} \selectfont This paper is accepted for presentation at IEEE Symposium Series on Computational Intelligence (IEEE SSCI 2017), Hawaii, USA, 2017.}
}

\begin{document}

\title{Opposition-based Ensemble\\ Micro-Differential Evolution}

\author{\IEEEauthorblockN{Hojjat Salehinejad\IEEEauthorrefmark{1},
Shahryar Rahnamayan\IEEEauthorrefmark{2}, and Hamid R. Tizhoosh\IEEEauthorrefmark{3}}
\IEEEauthorblockA{\IEEEauthorrefmark{1}Department of Electrical \& Computer Engineering, University of Toronto, Toronto, Canada}
\IEEEauthorblockA{\IEEEauthorrefmark{2}Department of Electrical \& Computer Engineering, University of Ontario Institute of Technology, Oshawa, Canada}
\IEEEauthorblockA{\IEEEauthorrefmark{3}KIMIA Lab, University of Waterloo, Waterloo, Canada\\
hojjat.salehinejad@mail.utoronto.ca, shahryar.rahnamayan@uoit.ca, tizhoosh@uwaterloo.ca}}

\maketitle
\thispagestyle{pageStyleOne}

\begin{abstract}
Differential evolution (DE) algorithm with a small population size is called Micro-DE (MDE). A small population size decreases the computational complexity but also reduces the exploration ability of DE by limiting the population diversity. In this paper, we propose the idea of combining ensemble mutation scheme selection and opposition-based learning concepts to enhance the diversity of population in MDE at mutation and selection stages. The proposed algorithm enhances the diversity of population by generating a random mutation scale factor per individual and per dimension, randomly assigning a mutation scheme to each individual in each generation, and diversifying individuals selection using opposition-based learning. This approach is easy to implement and does not require the setting of mutation scheme selection and mutation scale factor. Experimental results are conducted for a variety of objective functions with low and high dimensionality on the CEC Black-Box Optimization Benchmarking 2015 (CEC-BBOB 2015). The results show superior performance of the proposed algorithm compared to the other micro-DE algorithms.  
\end{abstract}


\IEEEpeerreviewmaketitle

\section{Introduction}

Population size plays an important role in performance of the population-based algorithms such as ant colony optimization \cite{salehinejad2010dynamic} and differential evolution (DE) \cite{salehinejad2017micro}. DE with a small population size is called micro-DE (MDE) \cite{salehinejad2014micro}. A small population size decreases the exploration ability of DE. The degrading performance is mainly due to the generation of fewer candidate individuals by the mutation scheme. This is while a large population size provides higher exploration ability, at the cost of more computational time. 

The lack of diversity in the population of MDE to explore the problem landscape can cause stagnation or premature convergence problems. The stagnation problem refers to a diverse and scattered population that cannot converge to a local optimum \cite{salehinejad2014micro}. From the other side, the lack of diversity in generating new individuals causes trapping of individuals in local optimums, generally referred to as premature convergence. A trade-off between the population size and evolution method of population improves the performance of algorithm \cite{salehinejad2016exploration}.

The standard version of DE uses a fixed value for mutation scale factor and a fixed mutation scheme. The MDE algorithm mostly suffers from lack of diversity in the population to generate new individuals \cite{salehinejad2017micro}. By vectorized randomizing the mutation scale factor (called MDEVM), which is a random mutation scale factor for each dimension of each individual, the micro-DE performs much better than utilizing a static mutation scale factor \cite{salehinejad2014micro}. Methods such as opposition-based DE (ODE) \cite{rahnamayan2008opposition} and centroid DE \cite{rahnamayan2014computing, rahnamayan2014centroid, salehinejad2016effects} have solely improved the performance of DE algorithm by proper diversifying the population in DE algorithm. These methods can also increase the performance of MDE.

In this paper, we enhance the MDEVM algorithm \cite{salehinejad2017micro,salehinejad2014micro,salehinejad20143d,salehinejad2014micro2}, by utilizing the idea of ensemble mutation scheme and opposition-based learning (OBL) \cite{tizhoosh2005opposition} to diversify the population. OBL generates the opposite of the current individuals (candidate solutions) and selects the best individuals with respect to objective function values \cite{salehinejad2014type}. At the mutation stage, each individual randomly selects a mutation scheme at each generation, called ensemble mutation scheme \cite{salehinejad2016exploration}. Combination of vectorized randomized mutation scale factor, ensemble mutation scheme, and OBL, has highly enhanced the performance of MDE. We call this method opposition-based ensemble micro-DE (OEMDE). 

\section{Related Works}
\label{sec:rw}
The main control parameters of DE algorithm are population size, mutation scale factor, crossover rate, and mutation scheme. Proper selection of these parameters helps to improve the performance of MDE.

\subsection{Mutation Scale Factor}

DE is about the difference of two or more vectors. Amplification of the difference vector is adjustable by the mutation scale factor $F$ which controls the evolving rate of the population. A small mutation scale factor decreases the exploration and may result in premature convergence. This is while a large mutation scale factor increases the exploration which results in a longer convergence time. The optimal value of $F$ is usually set based on the nature of the problem and experimental observations \cite{feoktistov2006differential}. MDEVM algorithm \cite{salehinejad2014micro} relaxes static selection of $F$ as a hyper-parameter by generating a random $F\in(0.1,1.5)$ for each dimension of an individual in the population \cite{salehinejad2014micro}. This randomness adds diversity to the population by amplifying the differences vectors in various scales. The experimental results in \cite{salehinejad2017micro} have demonstrated outstanding performance of this technique for diversity enhancement.

\subsection{Mutation Scheme and Crossover Rate}

The recommended range for the crossover rate is [0.3,0.9] in the literature \cite{gamperle2002parameter}. However, the optimal value is problem dependent. 
Aadaptive parameters can change as the evolution proceeds, such as self-adaptive DE (SDE) \cite{salman2007empirical} and memetic algorithm.

It is possible to use a pool of mutation strategies and a pool of values for each control parameter of the algorithm \cite{mallipeddi2011differential}. Ensemble algorithms use a pool of mutation schemes with two members, DE/rand/1 and DE/best/1. The strategy is to combine random parameters setting and mutation schemes to enhance convergence rate while maintaining the high diversity of population \cite{yu2015ensemble}. The proposed method in \cite{das2015switched} switches the scale factor and crossover rate in a uniform-random strategy for each individual of the population. The switching is performed within a particular range of the variables. A pool of mutation schemes is also utilized and the individuals use either the DE/rand/1 or DE/best/1 scheme. 
In order to implement the DE on systems with limited resources for real-time applications, such as embedded systems, a version of DE algorithm, called compact DE (cDE), is proposed \cite{mallipeddi2011ensemble}, \cite{mininno2011compact}. This algorithm uses probabilistic representation of the population, instead of the whole population for optimization. 
The logic behind cDE is producing candidates by sampling an evolving distribution in every generation. However, cDE algorithm suffers from premature convergence problem, particularly for high dimensional problems. This refers to the shrinkage of the probability function which models the population \cite{mallipeddi2011ensemble}. 
An ensemble version of cDE is proposed in \cite{mallipeddi2011ensemble}.

The memetic algorithms are hybrid evolutionary algorithms. These algorithms can solve optimization problems by utilizing deterministic local search within the evolutionary process. This technique along with the cDE algorithm is used to develop optimization algorithms on control cards \cite{neri2010memetic}. The objective is to design an optimization algorithm that can function in absence of full power computational systems.

A novel type of ensemble DE, called MPEDE, is proposed in \cite{wu2016differential}. It is a multi-population based approach which deploys a dynamic ensemble of multiple mutation strategies. It controls parameters of the algorithms such as mutation scale factor and crossover rate \cite{zhang2009jade}. This method uses the ``current-to-pbest/1", ``current-to-rand/1", and ``rand/1" mutation schemes \cite{wu2016differential}. Another method for unconstrained continuous optimization problems is $\mu$JADE \cite{brown2015mu}. It is an adaptive DE with a small population size approach. The $\mu$JADE uses a new mutation operator, called current-by-rand-to-pbest \cite{brown2015mu}.

Random perturbation method utilizes the mutation idea from the genetic algorithm (GA) to randomly change the population vector parameters with a fixes probability \cite{fajfar2012towards}. A modified DE with smaller population size can add disturbance to the mutation operation \cite{ren2010differential}. An adaptive system controls the intensity of disturbance based on the performance improvement during generations. 
A combination of modified Breeder GA mutation scheme and a random mutation scheme can help DE to avoid stagnation and/or premature convergence \cite{mohamed2012alternative}.

\subsection{Selection Scheme and Population Size Adaptation}

Large population size in DE algorithm causes high computational cost and adds more exploration ability to the population. The small population size in MDE algorithm reduces the number of function evaluations (per generation) with the cost of less exploration capability and a higher risk of premature convergence \cite{salehinejad2014micro}.

A population size adaptation method is proposed in \cite{yang2013improved} 
which measures the Euclidean distance between individuals and finds out if the population diversity is poor and/or it moves toward stagnation. Depending on the search situation, it generates new individuals. A population size reduction method for cDE is proposed in \cite{iacca2011super}, where the population size gradually reduces during the evolution. Cumu-DE is an adaptive DE and  the effective population size is adapted automatically using mechanism based on a probabilistic model \cite{aalto2015population}. The term effective refers to the effective part of the population where its size shrinks as the algorithm has more successful trials. Gradual reduction of the population size is another approach that has demonstrated higher robustness and efficiency compared to generic DE \cite{brest2008population}. 

OBL has Type-I and Type-II schemes \cite{tizhoosh2005opposition}, \cite{salehinejad2014type}. The Type-I scheme has enhanced the  performance of micro-ODE for image thresholding \cite{rahnamayan2008image}. This approach showed a better performance compared to the MDE algorithm. 

Small-size cooperative sub-populations are capable of finding sub-components of the original problem concurrently \cite{parsopoulos2009cooperative}. Combination of sub-components through their cooperation constructs a complete solution for the problem \cite{parsopoulos2009cooperative}. A MDE version of this method is proposed to evolve an indirect representation of the bin packing problem \cite{sotelo2013evolving}. The idea of self-adaptive population size is carried out to test absolute and relative encoding methods for DE \cite{teng2009self}. The reported simulation results on 20 benchmark problems denote the self-adaptive population size with relative encoding outperforms the absolute encoding method and the DE algorithm \cite{teng2009self}.

\section{Differential Evolution}
In general, during a black-box optimization process, an optimizer does not have a priori knowledge about the problem landscape. Therefore, optimizers normally start with randomly generated candidate solutions and then they try to get intelligence about the problem landscape during the optimization procedure. The DE algorithm, similar to other population-based algorithms, uses random initial vectors with uniform distribution at the beginning, called initial population. Through a number of generations, it tries to improve the initial candidate individuals toward an optimal solution. The population $\textbf{\textbf{P}}=\{\textbf{X}_{1},...,\textbf{X}_{N_{P}}\}$ has $N_{P}$ vectors in generation $g$, where $\textbf{X}_{i}$ is a $D$-dimensional vector defined as $\textbf{X}_{i}=(x_{i,1},...,x_{i,D})$. The classic version of DE algorithm is consisted of the following three major evolutionary operations: mutation, crossover, and selection.

\textit{Mutation:} This step selects three vectors randomly from the population such that $i_{1} \neq i_{2}\neq i_{3} \neq i$, where $i\in\{1,...,N_{P}\}$ s.t. $N_{P}\ge 4$ for each vector $\textbf{X}_{i}$. The mutant vector is calculated as
\begin{equation}
\textbf{V}_{i}=\textbf{X}_{i_{1}}+F(\textbf{X}_{i_{2}}-\textbf{X}_{i_{3}}),
\end{equation}  
where the mutation factor $F\in(0.1,1.5)$ is a real constant number that controls the amplification of the added differential variation of $(\textbf{X}_{i_{2}}-\textbf{X}_{i_{3}})$. The exploration of DE increases by selecting higher values for $F$.
 
\textit{Crossover:} The crossover operation increases diversity of the population by shuffling the mutant and parent vectors as follows
\begin{equation}
U_{i,d}= \left\{ \begin{array}{ll} V_{i,d}, & \textrm{rand}_{d}(0,1)\le Cr\: or\: d_{\textrm{rand}}=d\\
x_{i,d}, & \textrm{otherwise}
\end{array},\right.
\end{equation}
where $d=[1,...,D]$, $Cr\in[0,1]$ is the crossover rate control parameter, and $rand(a,b)$ generates a random number in the interval $[a,b]$ with a uniform distribution. Therefore, the trial vector $\textbf{U}_{i} \; \forall i \in\{1,...,N_{P}\}$ can be generated as
\begin{equation}
\textbf{U}_{i}=(U_{i,1},...,U_{i,D}).
\end{equation}
\textit{Selection:} The $\textbf{U}_{i}$ and $\textbf{X}_{i}$ vectors are evaluated and compared with respect to their fitness values; The one with better fitness is selected for the next generations.

\section{Opposition-Based Learning}
The typical idea in population-based algorithms is to generate some random initial candidates and then try to
guide them toward an optimal solution \cite{salehinejad2014type}. By considering candidates and corresponding opposite
candidates simultaneously, the search algorithm can have a more exploration coverage that increases the chance of
finding fitter candidates. This idea can be supported by the
probability theory, where for a specific problem, the opposite
of the current candidate solution has a higher chance to
be closer to the solution than a second random number \cite{wang2016using}, \cite{liu2014improved}. This
phenomenon can accelerate convergence rate of the optimizer. The concept of OBL can be categorized into Type-I
and Type-II schemes \cite{tizhoosh2005opposition}, \cite{salehinejad2014type}.

The Type-I OBL deals with
the relationship among candidate solutions, based on their attributes,
without considering the actual objective landscape. Generally
speaking, the Type-I OBL computation is easy due to its linear definition in the variable space. Type-I OBL can be considered as an approximation
of the genuine intellection of opposite computing, which is
the Type-II OBL. Type-II opposition scheme requires a profound understanding of the objective space. This concept
is typically difficult to be utilized in real-world applications,
since for the Type-I approach the variable space is already
known, while in the Type-II scenario a priori knowledge
about the landscape is required, which may not be available for black-box problems.

\begin{figure}[t]
\centerline{\includegraphics[width=3in]{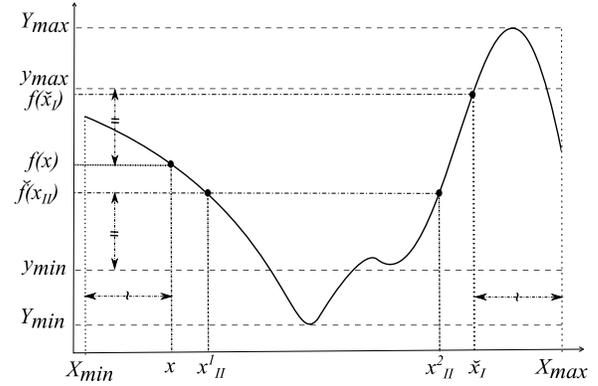}}
\caption{Type-I versus Type-II opposition definition for a
sample landscape. An arbitrary point $x$ with its corresponding objective value $f(x)$ is selected. The opposite of $x$ using Type-I approach is calculated using Eq.(\ref{eq:typei}) and denoted by $\breve{x}_{I}$, where $f(\breve{x}_{I})$ demonstrates its corresponding value on the landscape, objective value. By having $f(x)$, its Type-II opposite $\breve{f}(x_{II})$ is calculated using Eq.(\ref{eq:typeii}). The corresponding values of $\breve{f}(x_{II})$ in the variable space are denoted on the X-axis as $x_{II}^{1}$ and $x_{II}^{2}$. }  
\label{fig:obl}
\end{figure}

Figure~\ref{fig:obl} shows the Type-I and Type-II OBL schemes on a one dimensional problem landscape. OBL was initially defined with the Type-I trend with a min-max
oppositional computing scheme. This approach has received more attention due to
availability of variables' boundaries. The other method is centroid
opposition computing which works based on the centroid point of
population and does not require variables' boundaries.
The min-max Type-I OBL is defined for a $D$-dimensional point $x_{d}\in[X_{min,d},X_{max,d}]$ as
\begin{equation}
\breve{x}_{d}=X_{min,d}+X_{max,d}-x_{d}
\label{eq:typei}
\end{equation}
where $\breve{x}_{d}$ is the opposite of $x_{d}$ for $d=[1,...,D]$. By considering $\textbf{X}={(x_{1},...,x_{D})}$ as a point in the $D$-dimensional space and $\breve{\textbf{X}}={(\breve{x}_{1},...,\breve{x}_{D})}$ as the opposite point of $\textbf{X}$, if $f(\textbf{X})\leq f(\breve{\textbf{X}})$, where $f(\cdot)$ is the fitness value, $\breve{\textbf{X}}$ is considered as the better solution for a maximization problem.
In order to have a general definition, if $\psi\in\Psi$ in a $D$-dimensional space and $\Phi:\Psi\to\Psi$ is a one-to-one mapping function, which defines an oppositional relationship between two unique elements $\psi_{1}$ and $\psi_{2}$ of $\Psi$, the opposition concept for OBL Type-I is $\breve{\psi}=\Phi(\psi)$. In addition, if $\Phi(\psi_{1})=\psi_{2}$ and $\Phi(\psi_{2})=\psi_{1}$, the oppositional relationship is symmetric in this case. These approaches are discussed in \cite{rahnamayan2008opposition} and \cite{salehinejad2014type}.

For the point $\textbf{X}$ in Type-II OBL we have $f(\textbf{X})\in[Y_{min},Y_{max}]$ \cite{rahnamayan2012intuitive}, \cite{rahnamayan2008opposition}. Therefore, the min-max-based opposition for Type-II OBL is defined as
\begin{equation}
\breve{f}(\textbf{X})=Y_{min}+Y_{max}-f(\textbf{X}).
\label{eq:typeii}
\end{equation}
In practice, since the landscape boundaries are unknown, the boundaries $Y_{min}$ and $Y_{max}$ can be estimated using sampling as $y_{min}$ and $y_{max}$, respectively. The other approach can be using the centroid of known limits. However, it is proved in \cite{salehinejad2014type} that the centroid-based method can demonstrate better estimation compared to min-max-based opposition.

\alglanguage{pseudocode}
\begin{algorithm}[!ht]
\footnotesize
\textbf{Algorithm 1: Opposition-based Ensemble Micro-Differential Evolution (OEMDE)}
\begin{algorithmic}[1]
\State \textbf{Procedure} OEMDE
\State $g=0$ \textit{//} generation counter

\textit{//\textbf{Initial Population Generation}}

\For {$i=1 \to N_{P}$}
\For {$d=1 \to D$}
\State $\textbf{X}_{i,d}=x_{d}^{min}+rand(0,1) \times (x_{d}^{max}-x_{d}^{min})$
\EndFor
\EndFor

\textit{//\textbf{Compute opposite of generated population}}
\For {$i=1 \to N_{P}$}
\For {$d=1 \to D$}
\State $\breve{\textbf{X}}_{i,d}= x_{d}^{max}+x_{d}^{min}-\textbf{X}_{i,d}$
\EndFor
\EndFor

\textit{// \textbf{End of compute opposite of generated population}}
\State $S = \{min\{f(\breve{\textbf{X}}) \cup f(\textbf{X})\}:|S|=N_{P}\} $

\State $\textbf{P}^{g}=\{\textbf{X}_{i}|f(\textbf{X}_{i})\in S\}$ // \textbf{P} is the population 

\textit{// \textbf{End of Initial Population Generation}}

\While {($|BVF-VTR|>EVTR$ \& $NFC<NFC_{Max}$)}
\For {$i=1 \to N_{P}$}

\textit{// \textbf{Mutation}}

     \State \textit{Select J (number of individuals in the chosen mutation scheme) individuals from} $\textbf{P}^{g}$ \textit{where} $(i_{1} \neq i_{2}\neq i_{3} \neq i)$
     \For {$d=1 \to D$}
     \State $F_{i,d}=rand(0.1,1.5)$ 
     
     \State$\textbf{V}_{i,d}=f_{i}(F_{i,d},\textbf{X}_{i_{1},d},...,\textbf{X}_{i_{J},d})$
     \EndFor
     
     \textit{// \textbf{End of Mutation}}
     
     \textit{// \textbf{Crossover}}
     
         \For {$d=1 \to D$}
            \If {$rand(0,1)<Cr$ or $d_{rand}=d$}
                   \State $U_{i,d}=V_{i,d}$
                   \Else
                   \State $U_{i,d}=x_{i,d}$      
            \EndIf
          \EndFor
           
\textit{// \textbf{End of Crossover}}  

\textit{// \textbf{Selection}}         
           
          \If {$f(\textbf{U}_{i}) \leq f(\textbf{X}_{i})$}
             \State $\textbf{X}^{\prime}_{i}=\textbf{U}_{i}$
          \Else
             \State  $\textbf{X}^{\prime}_{i}=\textbf{X}_{i}$
          \EndIf   
          
\textit{// \textbf{End of Selection}}          
          
  \EndFor
  \State $\textbf{X}_{i}={\textbf{X}}^{\prime}_{i}$, $\forall i\in\{1,...,N_{P}\}$

\textit{//\textbf{Compute opposite of population}}
\For {$i=1 \to N_{P}$}
\For {$d=1 \to D$}
\State $\breve{\textbf{X}}_{i,d}= x_{d}^{max}+x_{d}^{min}-\textbf{X}_{i,d}$
\EndFor
\EndFor

\textit{// \textbf{End of compute opposite of generated population}}
\State $S = \{min\{f(\textbf{\v{X}}) \cup f(\textbf{X})\}:|S|=N_{P}\} $

  \State $g=g+1$
\State $\textbf{P}^{g}=\{\textbf{X}_{i}|f(\textbf{X}_{i})\in S\}$ // \textbf{P} is the population 

\EndWhile
\label{alg:OEMDE}
\end{algorithmic}

\end{algorithm}

\alglanguage{pseudocode}

\section{Proposed Algorithm}

In the classical DE, the mutation scale factor is set to a constant value and the mutation scheme is identical for all individuals during generations. In the proposed OEMDE algorithm, we enhance the exploration ability of MDE by proposing a combination of three strategies: 1) vectorized randomization of mutation scale factor; 2) utilization of ensemble mutation scheme; 3) population initialization and generation jumping based on the opposition-based learning. 

The OEMDE algorithm is presented in Algorithm 1, where \textit{BVF} is the best value so far, \textit{VTR} is the value to reach, \textit{EVTR} is error value to reach, \textit{NFC} is number of function calls, and \textit{$NFC_{Max}$} is the maximum number of function calls. 

\subsection{Vectorized Random Mutation Factor}
The mutation scale factor proposed in \cite{salehinejad2014micro} is generated in a vectorized random scheme such that it is a different value for 
each dimension of an individual. Therefore, the mutation factor can be defined for each individual $i$ as
\begin{equation}
\textbf{F}_{i}=(F_{i,1},...,F_{i,D}),\: \forall i \in {1,...,N_{P}},
\end{equation}  
where $F_{i,d}=rand(0.1,1.5)$ \cite{salehinejad2014micro}.

\subsection{Ensemble Mutation Scheme}
The OEMDE algorithm randomly assigns a mutation scheme to each individual $i$ of the population from a pool of mutation schemes, Table~\ref{T:MSs}. The mutation scheme is a function of mutation scale factor and $J$ selected parents for each individual such as
\begin{equation}
\textbf{V}_{i,d}=f_{i}(F_{i,d},\textbf{X}_{i_{1},d},...,\textbf{X}_{i_{J},d})
\end{equation}
This function adds diversity into the population by using vectorized random mutation scale factor as well as using a random mutation scheme for each individual in each generation. Random selection of mutation scheme from a uniform distribution results in enhancing the diversity in the population and therefore, more exploration capability of the DE algorithm. Diversity in mutation scheme selection is particularly important for MDE algorithm, where reducing the population size to a micro-level results in a loss of diversity in the population. This is one of the main causes of premature convergence and stagnation.

\begin{table}[t]
\captionsetup{font=footnotesize}
\footnotesize
\caption{Pool of mutation schemes.}
\begin{center}
\begin{adjustbox}{width=0.49\textwidth}
\begin{tabular}{|c|c|}
\hline
\multirow{1}{*}{Mutation Scheme}&\multicolumn{1}{c|}{Equation}\\
\hhline{--}
\hline
\hline
DE/rand/1&$\mathbf{V}_{i}=\textbf{X}_{i_{1}}+F(\textbf{X}_{i_{2}}-\textbf{X}_{i_{3}})$  \\ \hline
DE/best/1&$\mathbf{V}_{i}=\textbf{X}_{{best}}+F(\textbf{X}_{i_{1}}-\textbf{X}_{i_{2}})$ \\ \hline
DE/target-to-best/1&$\mathbf{V}_{i}=\textbf{X}_{i}+F(\textbf{X}_{{best}}-\textbf{X}_{i})+F(\textbf{X}_{i_{1}}-\textbf{X}_{i_{2}})$ \\ \hline
 DE/rand/2&$\mathbf{V}_{i}=\textbf{X}_{i_{1}}+F(\textbf{X}_{i_{2}}-\textbf{X}_{i_{3}})+F(\textbf{X}_{i_{4}}-\textbf{X}_{i_{5}})$ \\ \hline
DE/best/2&$\mathbf{V}_{i}=\textbf{X}_{{best}}+F(\textbf{X}_{i_{1}}-\textbf{X}_{i_{2}})+F(\textbf{X}_{i_{3}}-\textbf{X}_{i_{4}})$ \\ \hline
\end{tabular}
\label{T:MSs}

\end{adjustbox}
\end{center}
\end{table}

\subsection{Opposition-based Learning}
The OBL not only considers the current population, but also gives a chance to the opposite individuals of the population to cooperate in the evolution. We use OBL in population initialization as well as for generation jumping similar to ODE algorithm \cite{rahnamayan2008opposition}.  
As Algorithm~1 shows, the opposite of individual $i$ at dimension $d$ is computed as
\begin{equation}
\breve{\textbf{X}}_{i,d}= x_{d}^{max}+x_{d}^{min}-\textbf{X}_{i,d}
\end{equation}
where $x_{d}^{max}$ and $x_{d}^{min}$ are the maximum and minimum boundaries of dimension $d$, respectively. The objective function values of the opposite population $f(\breve{\textbf{X}})$ is computed and $N_{P}$ individuals with minimum objective function values are selected such as 
\begin{equation}
S = \{min\{f(\breve{\textbf{X}}) \cup f(\textbf{X})\}:|S|=N_{P}\} 
\end{equation}
and 
\begin{equation}
\textbf{P}^{g}=\{\textbf{X}_{i}|f(\textbf{X}_{i})\in S\},
\end{equation}
where $\textbf{P}^{g}$ is the selected population for generation $g$.

\begin{figure*}[!htbp]
\captionsetup{font=footnotesize}
\centering 
        \begin{subfigure}[b]{0.25\textwidth}
                \includegraphics[width=1.1\textwidth]{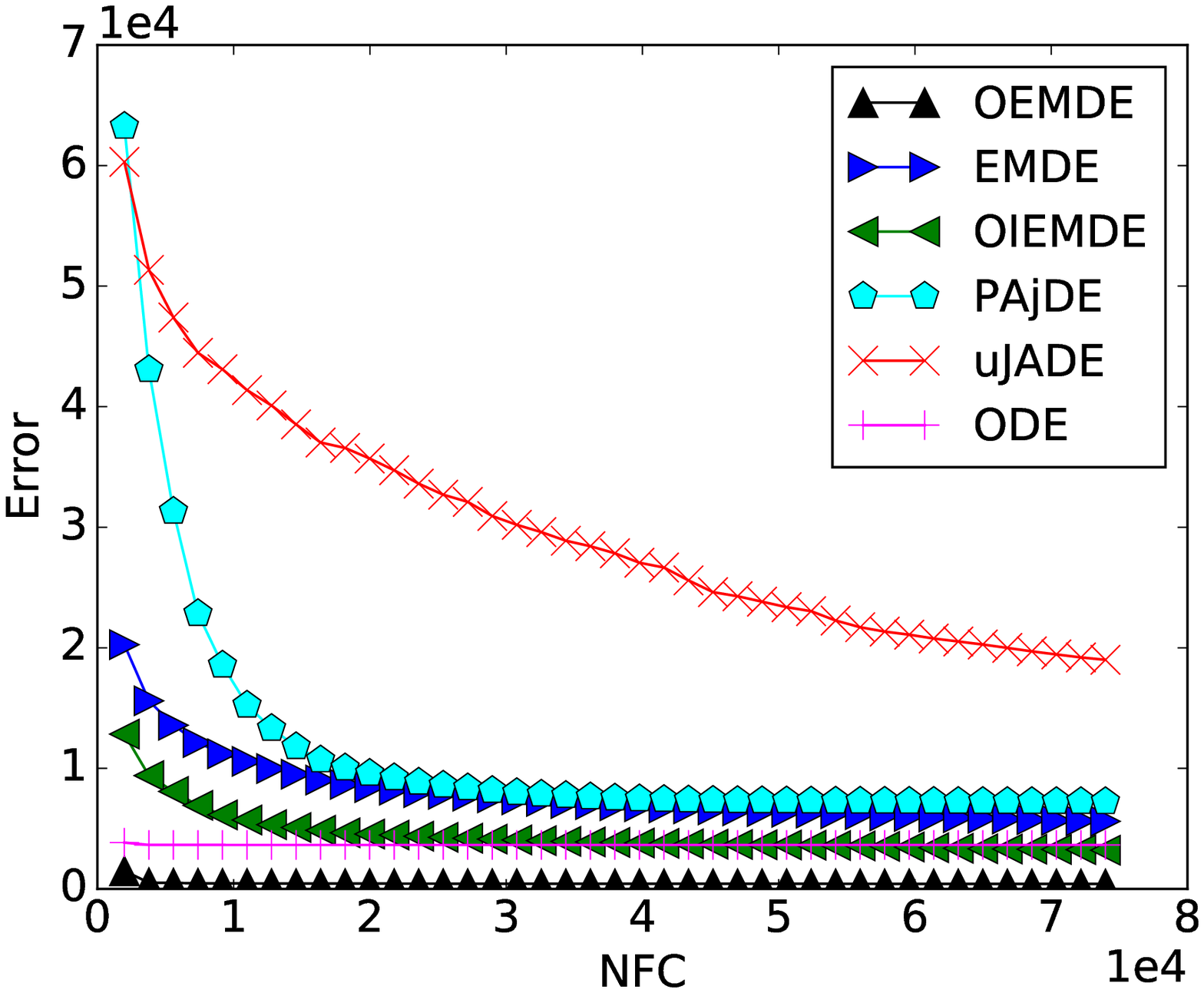}
                \caption{$f_{1}$.}
                \label{fig:}
                        \end{subfigure}%
        \begin{subfigure}[b]{0.25\textwidth}
                \includegraphics[width=1.1\textwidth]{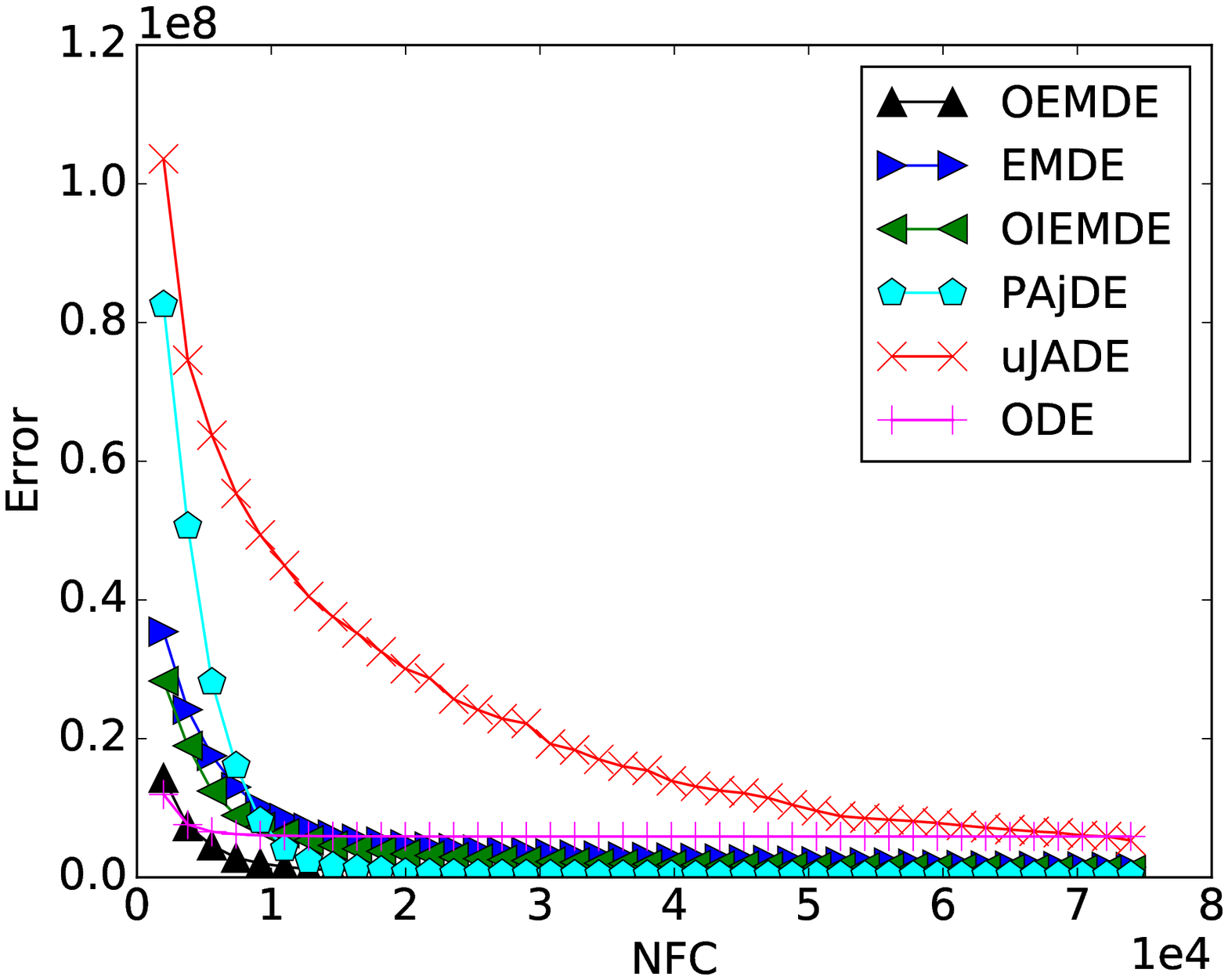}
                \caption{$f_{6}$.}
                \label{fig:}
        \end{subfigure}%
        \begin{subfigure}[b]{0.25\textwidth}
                \includegraphics[width=1.1\textwidth]{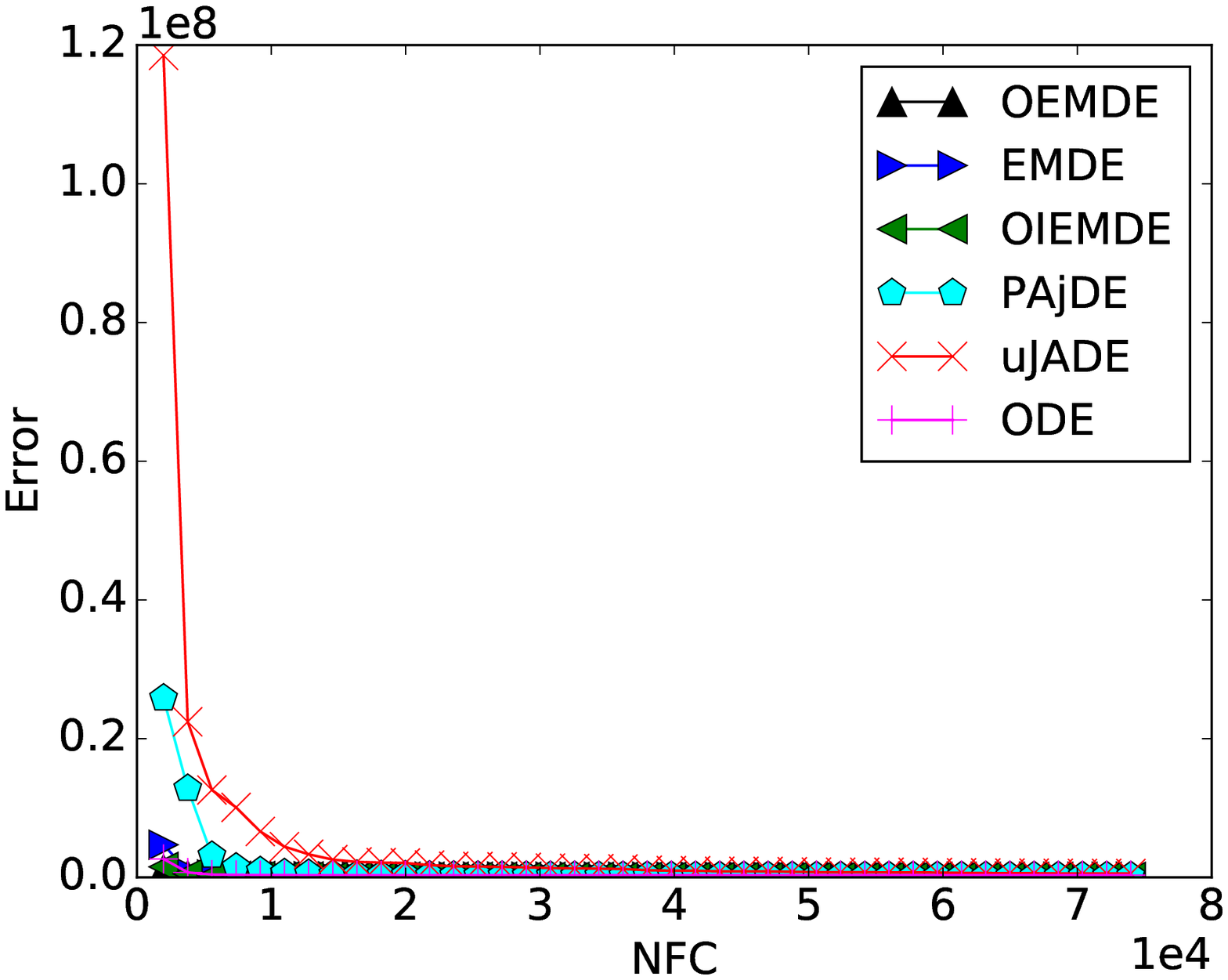}
                \caption{$f_{15}$.}
                \label{fig:}
        \end{subfigure}%
                \begin{subfigure}[b]{0.25\textwidth}
                \includegraphics[width=1.1\textwidth]{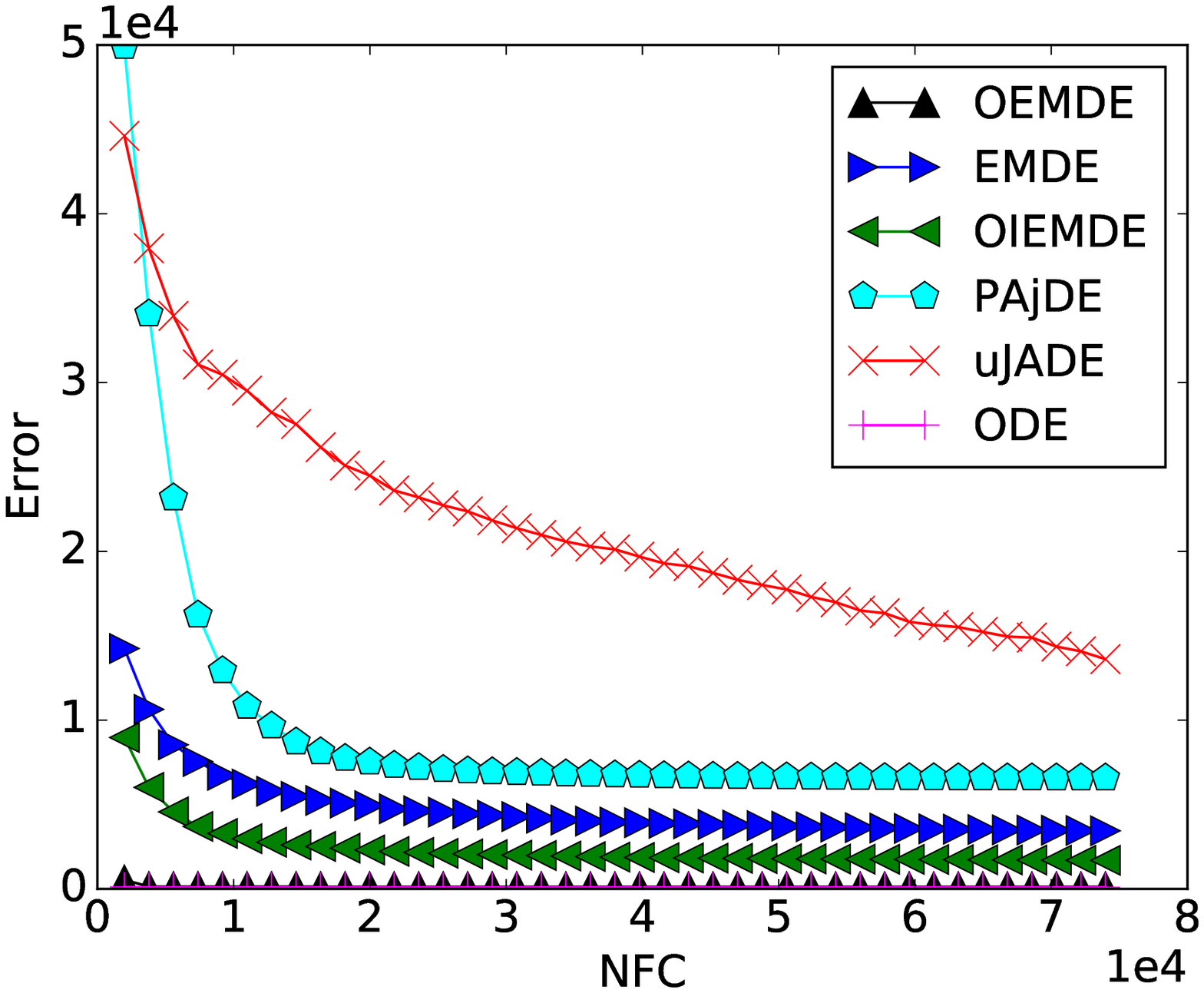}
                \caption{$f_{21}$.}
                \label{fig:}
        \end{subfigure}%
        \caption{Error versus number of function calls (NFC) for $N_{P}=6$ and $\textbf{D=30}$.}
        \label{fig:d30} 
\end{figure*}
\begin{figure*}[!htbp]
\captionsetup{font=footnotesize}
\centering 
        \begin{subfigure}[b]{0.25\textwidth}
                \includegraphics[width=1.1\textwidth]{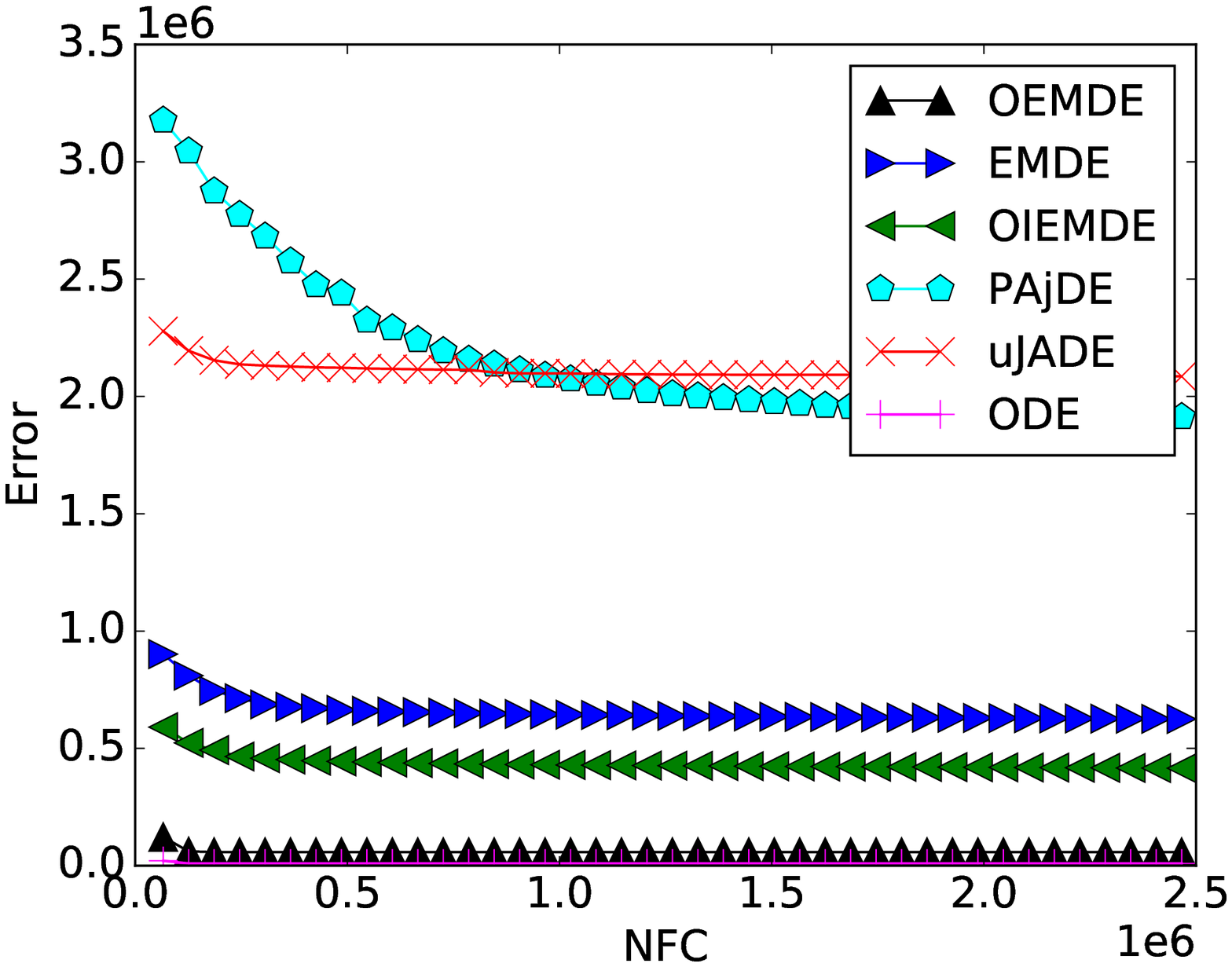}
                \caption{$f_{1}$.}
                \label{fig:}
        \end{subfigure}%
        \begin{subfigure}[b]{0.25\textwidth}
                \includegraphics[width=1.1\textwidth]{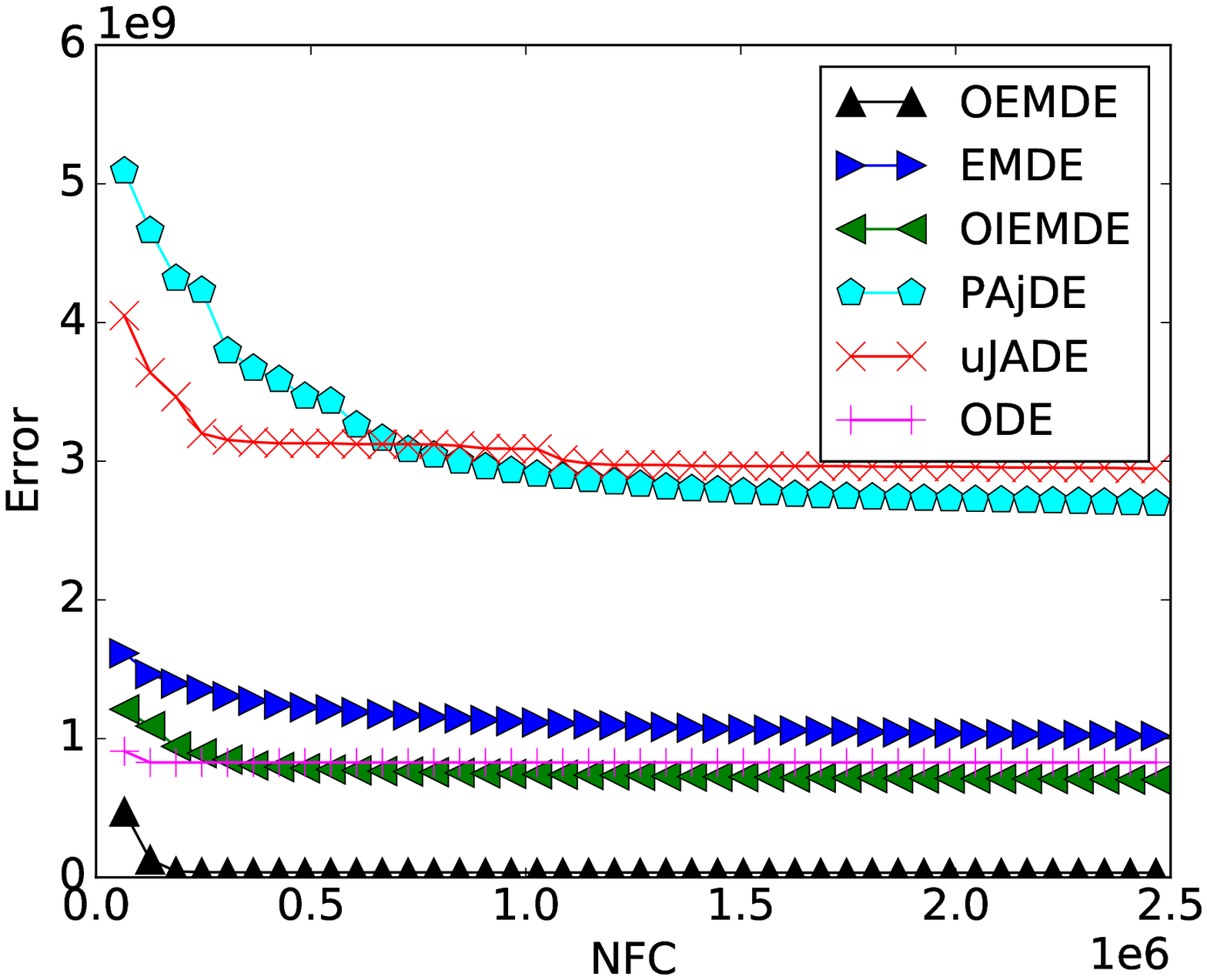}
                \caption{$f_{6}$.}
                \label{fig:}
        \end{subfigure}%
        \begin{subfigure}[b]{0.25\textwidth}
                \includegraphics[width=1.1\textwidth]{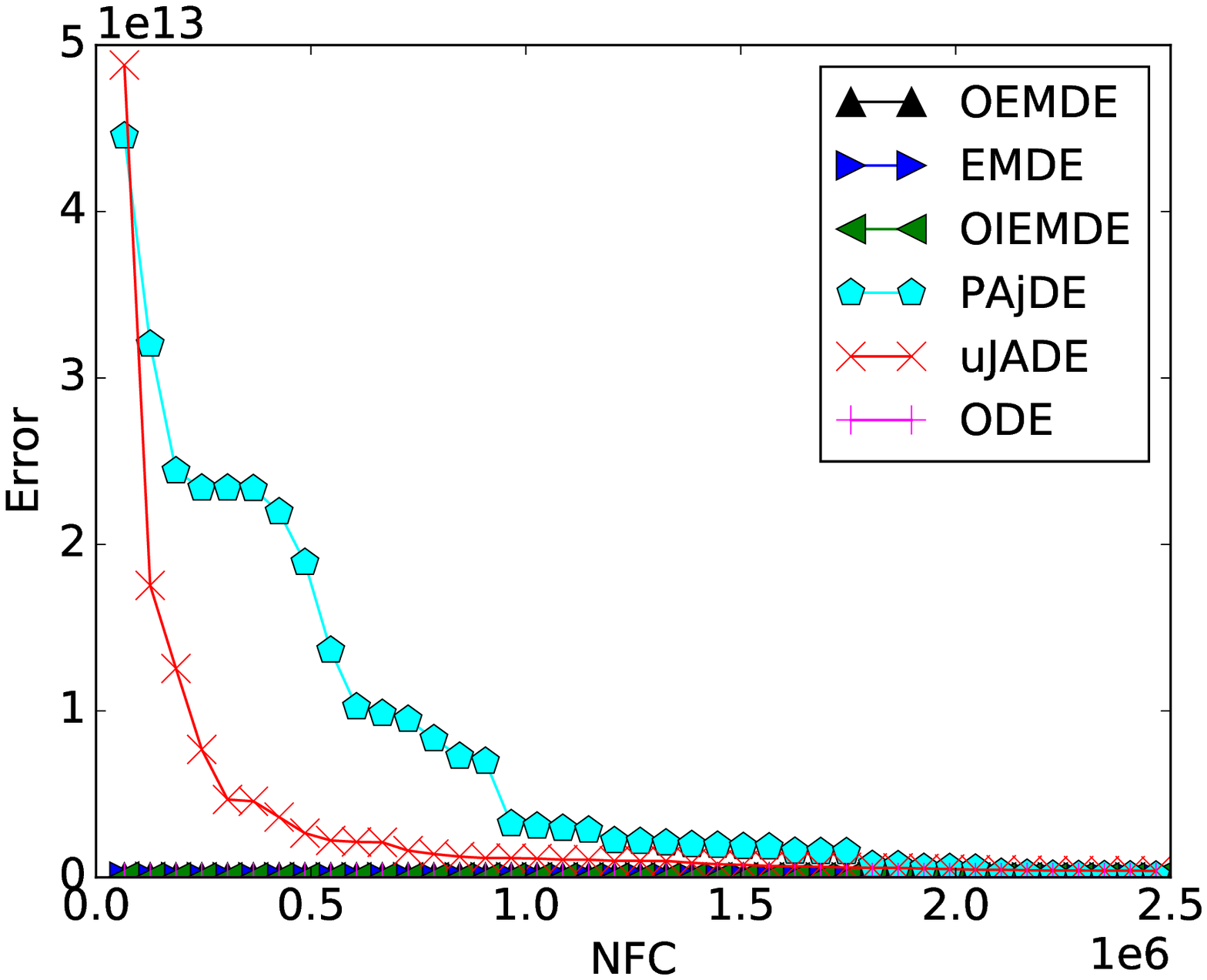}
                \caption{$f_{15}$.}
                \label{fig:}
        \end{subfigure}%
                \begin{subfigure}[b]{0.25\textwidth}
                \includegraphics[width=1.1\textwidth]{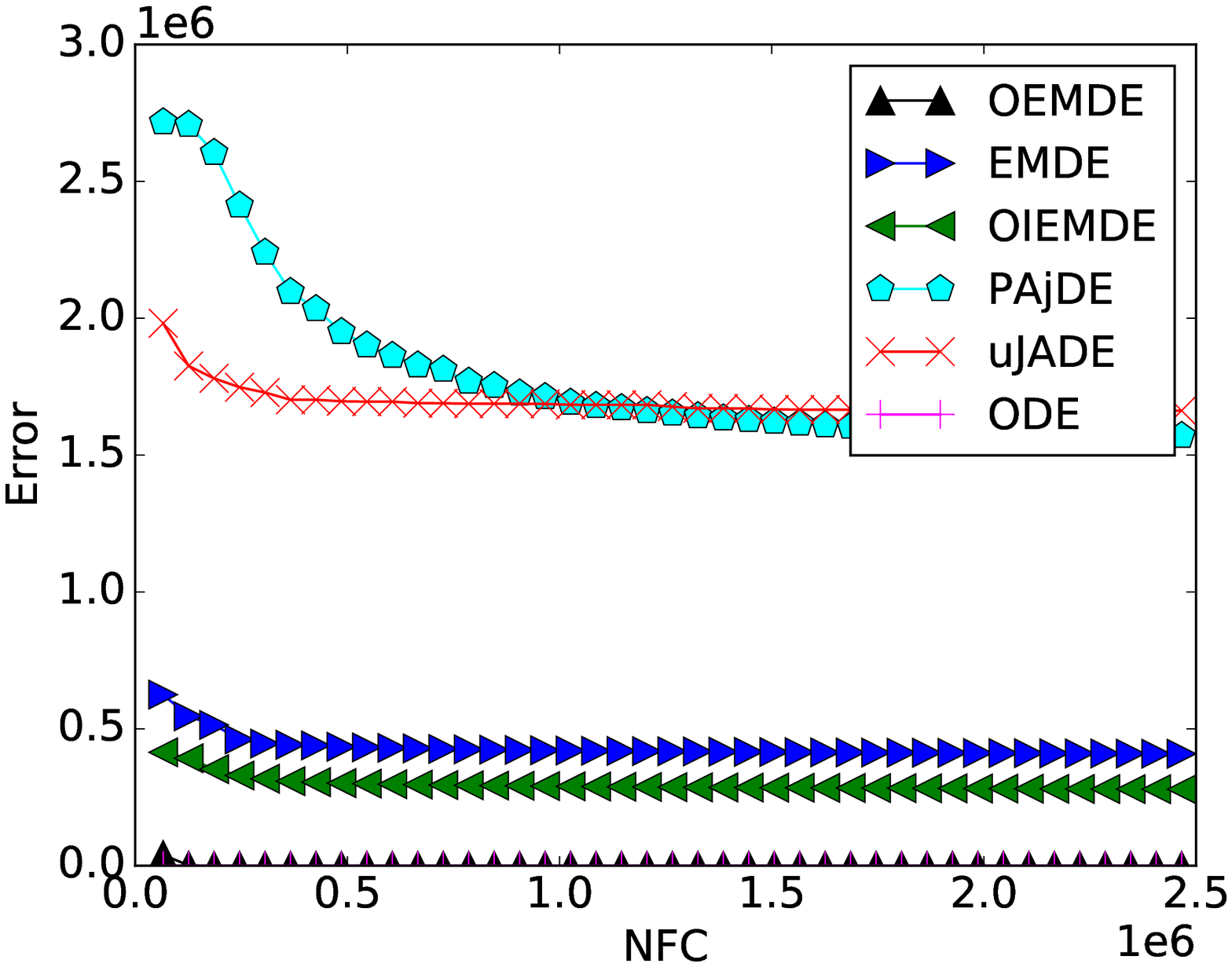}
                \caption{$f_{21}$.}
                \label{fig:}
        \end{subfigure}%
        \caption{Error versus number of function calls (NFC) for $N_{P}=6$ and $\textbf{D=1000}$.}
        \label{fig:d1000} 
\end{figure*}

\begin{table}[t]
\caption{Summary of statistical tests of OEMDE vs. EMDE, OIEMDE, PAjDE, $\mu$JADE, and ODE for $N_{P}=6$ and dimensions $D=\{10, 30, 50, 100, 500, 1000\}$.}
\begin{center}
\footnotesize
\begin{adjustbox}{width=0.49\textwidth}
\begin{tabular}{|c|c|c|c|c|c|c|c|c|c|c|c|c|c|c|c|}
\hline
\multirow{2}{*}{$D$}& \multicolumn{3}{|c|}{EMDE}& \multicolumn{3}{|c|}{OIEMDE}& \multicolumn{3}{|c|}{PAjDE}& \multicolumn{3}{|c|}{$\mu$JADE}& \multicolumn{3}{|c|}{ODE}\\ \hhline{~---------------}
\multirow{2}{*}{}& +&=&-&  +&=&-&  +&=&-&  +&=&-&   +&=&- \\ \hhline{----------------}
30	&	\textbf{14}	&	8	&	2	&	9	&	\textbf{12}	&	3	&	\textbf{19}	&	4	&	1	&	\textbf{18}	&	4	&	2							&	6	&	\textbf{17}	&	1	\\ \hline
50	&	7	&	\textbf{17}	&	0	&	\textbf{12}	&	11	&	1	&	\textbf{17}	&	5	&	2	&	\textbf{17}	&	6	&	1							&	\textbf{12}	&	6	&	6	\\ \hline
100	&	\textbf{18}	&	5	&	1	&	\textbf{13}	&	7	&	4	&	\textbf{21}	&	2	&	1	&	1	&	\textbf{22}	&	1							&	7	&	8	&	\textbf{9}	\\ \hline
200	&	\textbf{21}	&	2	&	1	&	\textbf{20}	&	2	&	2	&	\textbf{23}	&	1	&	0	&	\textbf{22}	&	2	&	0							&	6	&	\textbf{10}	&	8	\\ \hline
500	&	\textbf{21}	&	2	&	1	&	\textbf{18}	&	6	&	0	&	\textbf{22}	&	2	&	0	&	\textbf{22}	&	2	&	0							&	7	&	7	&	\textbf{10}	\\ \hline
1000	&	\textbf{21}	&	2	&	1	&	\textbf{18}	&	6	&	0	&	\textbf{22}	&	2	&	0	&	2	&	\textbf{22}	&	0							&	7	&	\textbf{10}	&	7	\\ \hline

\end{tabular}
\end{adjustbox}
\label{T:summary}
\end{center}
\vspace{-5mm}

\end{table}

\section{Experimental Results}

In this section, we discuss performance of the proposed algorithm versus the PAjDE \cite{yang2013improved}, $\mu$JADE \cite{brown2015mu}, ODE \cite{rahnamayan2008opposition}, ensemble micro differential evolution (EMDE) \cite{salehinejad2016exploration}, and EMDE with opposition in initialization (OIEMDE). 

\subsection{Benchmark Problems}
We use the black-box optimization benchmarking (CEC-BBOB-2015) functions given at CEC-2015 \cite{hansen2010real}. It has 24 noise-free functions divided into five classes which are separable functions~($f_{1}-f_{5}$),~functions with low or moderate conditioning ($f_{6}-f_{9}$), functions with high conditioning and uni-modal ($f_{10}-f_{14}$), multi-modal functions with adequate global structure ($f_{15}-f_{19}$), and multi-modal functions with weak global structure ($f_{20}-f_{24}$).

\subsection{Setting}
Experiments are conducted for six different dimensions $D\in\{30,50,100,200,500,1000\}$, crossover rate is set to $Cr=0.9$, maximum number of function calls is $NFC_{max} = 5000D$, and error-value-to-reach is set to $EVTR = 10^{-8}$. The mutation scheme pool is given as in Table~\ref{T:MSs}. Since one of the mutation schemes has 5 parents/individuals,  the minimum population size is set to $N_{P}=6$~\cite{salehinejad2017micro}. The experiments are conducted 30 independent times. Parameters setting of the algorithms compared with the OEMDE are set based on the best setting reported in their corresponding papers. Settings of OEIMDE is similar to OEMDE.

\begin{table*}[t]
\captionsetup{font=footnotesize}
\tiny
\caption{Error and standard deviation of the proposed OEMDE algorithm versus state of the art. The \textit{W} represents the Wilcoxon statistical test results compared to OEMDE. The signs `$+$' and `$-$' show significance of the OEMDE over the competitive algorithm and vice-versa, respectively; `$=$' shows neither is significant. $\textbf{D=50}$ and $N_{P}=6$.}
\vspace{-0.3cm}
\begin{center}
\begin{tabular}{|c|c|c|c|c|c|c|c|c|c|c|c|}
\hline
\multirow{1}{*}{$F$} & \multicolumn{1}{c|}{OEMDE}  & \multicolumn{2}{c|}{EMDE} &  \multicolumn{2}{c|}{OIEMDE} & \multicolumn{2}{c|}{PAjDE} & \multicolumn{2}{c|}{$\mu$JADE}& \multicolumn{2}{c|}{ODE}\\ \hhline{~-----------}
 & \multicolumn{1}{c|}{Error}  & \multicolumn{1}{c|}{Error} &\multicolumn{1}{c|}{W} &  \multicolumn{1}{c|}{Error} &\multicolumn{1}{c|}{W} &\multicolumn{1}{c|}{Error} &\multicolumn{1}{c|}{W} & \multicolumn{1}{c|}{Error} &\multicolumn{1}{c|}{W}& \multicolumn{1}{c|}{Error} &\multicolumn{1}{c|}{W}\\
\hline
\hline
1	&	6.00e+02$\pm$1.48e+03	&	1.09e+04$\pm$3.26e+03	&	$+$	&	7.11e+03$\pm$2.41e+03	&	$+$	&	2.27e+04$\pm$7.50e+03	&	$+$	&	3.48e+04$\pm$6.37e+03	&	$+$	&	3.26e+02$\pm$5.97e+02	&	$=$	\\ \hline
2	&	5.10e+07$\pm$5.90e+07	&	2.91e+08$\pm$1.61e+08	&	$+$	&	1.71e+08$\pm$7.35e+07	&	$+$	&	4.38e+08$\pm$2.41e+08	&	$+$	&	4.18e+08$\pm$1.31e+08	&	$+$	&	1.95e+07$\pm$6.83e+06	&	$-$	\\ \hline
3	&	1.20e+06$\pm$1.78e+06	&	3.61e+06$\pm$7.64e+06	&	$=$	&	1.03e+06$\pm$1.72e+06	&	$=$	&	4.65e+07$\pm$9.40e+07	&	$+$	&	1.81e+06$\pm$3.56e+06	&	$=$	&	3.56e+07$\pm$7.77e+07	&	$+$	\\ \hline
4	&	6.39e+05$\pm$1.21e+06	&	1.39e+06$\pm$4.87e+05	&	$+$	&	8.70e+05$\pm$4.49e+05	&	$=$	&	2.88e+06$\pm$1.04e+06	&	$+$	&	3.81e+06$\pm$6.00e+05	&	$+$	&	9.23e+05$\pm$2.13e+06	&	$+$	\\ \hline
5	&	4.84e+02$\pm$1.62e+02	&	6.45e+02$\pm$2.20e+02	&	$+$	&	4.63e+02$\pm$1.62e+02	&	$=$	&	5.03e+02$\pm$3.30e+02	&	$=$	&	1.18e+02$\pm$7.09e+01	&	$-$	&	1.20e+03$\pm$3.08e+02	&	$+$	\\ \hline
6	&	3.69e+06$\pm$4.39e+06	&	4.07e+06$\pm$5.83e+06	&	$=$	&	2.77e+06$\pm$3.05e+06	&	$=$	&	4.84e+05$\pm$6.61e+05	&	$-$	&	4.05e+06$\pm$5.50e+06	&	$=$	&	3.35e+07$\pm$6.65e+07	&	$+$	\\ \hline
7	&	1.06e+04$\pm$2.29e+04	&	6.99e+04$\pm$1.62e+04	&	$+$	&	4.58e+04$\pm$1.54e+04	&	$+$	&	1.42e+05$\pm$5.07e+04	&	$+$	&	2.45e+05$\pm$4.42e+04	&	$+$	&	3.60e+04$\pm$1.23e+05	&	$+$	\\ \hline
8	&	2.56e+07$\pm$8.21e+07	&	1.23e+09$\pm$4.60e+08	&	$+$	&	5.13e+08$\pm$3.29e+08	&	$+$	&	6.79e+09$\pm$3.61e+09	&	$+$	&	8.70e+09$\pm$3.75e+09	&	$+$	&	2.28e+05$\pm$4.15e+04	&	$-$	\\ \hline
9	&	8.56e+08$\pm$2.78e+09	&	8.11e+08$\pm$4.26e+08	&	$=$	&	2.77e+08$\pm$1.68e+08	&	$=$	&	2.87e+09$\pm$1.94e+09	&	$+$	&	9.44e+09$\pm$1.90e+09	&	$+$	&	1.29e+04$\pm$5.62e+04	&	$-$	\\ \hline
10	&	4.98e+08$\pm$5.36e+08	&	1.40e+08$\pm$6.23e+07	&	$-$	&	6.75e+07$\pm$3.59e+07	&	$-$	&	1.83e+08$\pm$9.13e+07	&	$-$	&	3.63e+08$\pm$9.71e+07	&	$=$	&	3.16e+08$\pm$1.25e+09	&	$=$	\\ \hline
11	&	1.57e+06$\pm$3.65e+06	&	1.09e+05$\pm$4.97e+04	&	$=$	&	1.59e+05$\pm$1.05e+05	&	$=$	&	2.74e+05$\pm$1.00e+05	&	$=$	&	1.34e+05$\pm$2.07e+04	&	$=$	&	5.30e+05$\pm$9.95e+05	&	$-$	\\ \hline
12	&	6.11e+11$\pm$1.89e+12	&	4.81e+11$\pm$8.43e+11	&	$=$	&	1.17e+12$\pm$3.86e+12	&	$=$	&	2.01e+13$\pm$8.50e+13	&	$=$	&	6.39e+14$\pm$1.64e+15	&	$=$	&	2.66e+15$\pm$7.40e+15	&	$+$	\\ \hline
13	&	2.54e+03$\pm$1.44e+03	&	1.96e+04$\pm$2.74e+03	&	$+$	&	1.54e+04$\pm$2.64e+03	&	$+$	&	2.86e+04$\pm$3.82e+03	&	$+$	&	3.82e+04$\pm$4.07e+03	&	$+$	&	3.03e+03$\pm$1.77e+02	&	$+$	\\ \hline
14	&	6.72e+03$\pm$1.28e+04	&	3.88e+03$\pm$1.45e+03	&	$=$	&	2.74e+03$\pm$1.31e+03	&	$=$	&	9.42e+03$\pm$6.73e+03	&	$=$	&	1.42e+04$\pm$4.03e+03	&	$+$	&	3.52e+01$\pm$9.10e+01	&	$-$	\\ \hline
15	&	6.79e+05$\pm$1.60e+06	&	2.81e+05$\pm$2.37e+05	&	$=$	&	1.36e+05$\pm$4.02e+04	&	$=$	&	6.91e+05$\pm$8.70e+05	&	$=$	&	1.21e+06$\pm$8.57e+05	&	$=$	&	2.23e+06$\pm$5.90e+06	&	$+$	\\ \hline
16	&	3.16e+02$\pm$5.34e+02	&	1.61e+03$\pm$5.49e+02	&	$+$	&	8.28e+02$\pm$3.97e+02	&	$+$	&	3.08e+03$\pm$1.38e+03	&	$+$	&	5.11e+03$\pm$1.14e+03	&	$+$	&	2.08e+02$\pm$1.65e+01	&	$=$	\\ \hline
17	&	6.85e+04$\pm$7.70e+04	&	1.44e+05$\pm$7.71e+04	&	$+$	&	1.07e+05$\pm$5.59e+04	&	$=$	&	5.33e+05$\pm$2.48e+05	&	$+$	&	9.86e+05$\pm$2.09e+05	&	$+$	&	4.28e+05$\pm$4.79e+05	&	$+$	\\ \hline
18	&	1.61e+05$\pm$1.32e+05	&	2.45e+05$\pm$1.09e+05	&	$+$	&	1.78e+05$\pm$1.09e+05	&	$=$	&	5.04e+05$\pm$1.88e+05	&	$+$	&	1.16e+06$\pm$3.25e+05	&	$+$	&	4.86e+05$\pm$5.89e+05	&	$+$	\\ \hline
19	&	2.64e+03$\pm$7.95e+03	&	3.90e+04$\pm$2.01e+04	&	$+$	&	1.19e+04$\pm$8.83e+03	&	$+$	&	1.46e+05$\pm$1.16e+05	&	$+$	&	4.42e+05$\pm$1.24e+05	&	$+$	&	9.62e+03$\pm$4.17e+04	&	$+$	\\ \hline
20	&	1.23e+06$\pm$4.49e+06	&	1.53e+07$\pm$4.81e+06	&	$+$	&	3.77e+06$\pm$0.00e+00	&	$=$	&	3.36e+07$\pm$1.35e+07	&	$+$	&	4.73e+07$\pm$1.15e+07	&	$+$	&	1.18e+07$\pm$4.57e+07	&	$+$	\\ \hline
21	&	4.29e+02$\pm$1.99e+02	&	8.62e+03$\pm$2.52e+03	&	$+$	&	4.39e+03$\pm$1.68e+03	&	$+$	&	1.52e+04$\pm$5.93e+03	&	$+$	&	2.65e+04$\pm$5.84e+03	&	$+$	&	3.87e+02$\pm$1.76e+00	&	$=$	\\ \hline
22	&	1.21e+02$\pm$1.65e+01	&	7.12e+03$\pm$2.67e+03	&	$+$	&	3.93e+03$\pm$1.80e+03	&	$+$	&	1.68e+04$\pm$5.91e+03	&	$+$	&	2.45e+04$\pm$4.86e+03	&	$+$	&	1.48e+02$\pm$1.06e+02	&	$=$	\\ \hline
23	&	6.00e+02$\pm$1.15e+03	&	7.60e+03$\pm$1.84e+03	&	$+$	&	4.55e+03$\pm$2.65e+03	&	$+$	&	1.94e+04$\pm$7.17e+03	&	$+$	&	2.77e+04$\pm$4.73e+03	&	$+$	&	2.18e+02$\pm$3.09e+00	&	$=$	\\ \hline
24	&	4.24e+06$\pm$1.28e+07	&	7.09e+07$\pm$2.10e+07	&	$+$	&	3.84e+07$\pm$1.75e+07	&	$+$	&	1.93e+08$\pm$5.85e+07	&	$+$	&	2.57e+08$\pm$6.42e+07	&	$+$	&	7.74e+02$\pm$7.08e+01	&	$-$	\\ \hline
\end{tabular}
\label{T:D50NP6}
\end{center}
\end{table*}
\begin{table*}[t]
\captionsetup{font=footnotesize}
\tiny
\caption{Error and standard deviation of the proposed OEMDE algorithm versus state of the art. The \textit{W} represents the Wilcoxon statistical test results compared to OEMDE. The signs `$+$' and `$-$' show significance of the OEMDE over the competitive algorithm and vice-versa, respectively; `$=$' shows neither is significant. $\textbf{D=100}$ and $N_{P}=6$.}
\vspace{-0.3cm}
\begin{center}
\begin{tabular}{|c|c|c|c|c|c|c|c|c|c|c|c|}
\hline
\multirow{1}{*}{$F$} & \multicolumn{1}{c|}{OEMDE}  & \multicolumn{2}{c|}{EMDE} &  \multicolumn{2}{c|}{OIEMDE} & \multicolumn{2}{c|}{PAjDE} & \multicolumn{2}{c|}{$\mu$JADE}& \multicolumn{2}{c|}{ODE}\\ \hhline{~-----------}
 & \multicolumn{1}{c|}{Error}  & \multicolumn{1}{c|}{Error} &\multicolumn{1}{c|}{W} &  \multicolumn{1}{c|}{Error} &\multicolumn{1}{c|}{W} &\multicolumn{1}{c|}{Error} &\multicolumn{1}{c|}{W} & \multicolumn{1}{c|}{Error} &\multicolumn{1}{c|}{W}& \multicolumn{1}{c|}{Error} &\multicolumn{1}{c|}{W}\\
\hline
\hline
1	&	1.49e+03$\pm$4.07e+03	&	3.24e+04$\pm$5.13e+03	&	$+$	&	2.16e+04$\pm$5.56e+03	&	$+$	&	7.42e+04$\pm$1.09e+04	&	$+$	&	1.25e+05$\pm$1.57e+04	&	$+$	&	2.07e+03$\pm$6.64e+03	&	$=$	\\ \hline
2	&	6.24e+07$\pm$9.33e+07	&	1.06e+09$\pm$3.88e+08	&	$+$	&	7.52e+08$\pm$3.39e+08	&	$+$	&	1.86e+09$\pm$6.06e+08	&	$+$	&	1.92e+09$\pm$5.09e+08	&	$+$	&	7.55e+08$\pm$3.16e+09	&	$+$	\\ \hline
3	&	1.73e+07$\pm$2.54e+07	&	2.07e+07$\pm$2.04e+07	&	$=$	&	7.93e+06$\pm$8.05e+06	&	$-$	&	7.39e+08$\pm$1.25e+09	&	$+$	&	8.09e+07$\pm$9.90e+07	&	$+$	&	2.92e+08$\pm$4.56e+08	&	$+$	\\ \hline
4	&	1.22e+06$\pm$2.30e+06	&	5.08e+06$\pm$1.17e+06	&	$+$	&	3.41e+06$\pm$1.13e+06	&	$=$	&	1.10e+07$\pm$2.79e+06	&	$+$	&	1.59e+07$\pm$1.86e+06	&	$+$	&	1.80e+06$\pm$7.65e+06	&	$=$	\\ \hline
5	&	1.56e+03$\pm$4.10e+02	&	1.96e+03$\pm$5.42e+02	&	$+$	&	1.59e+03$\pm$4.09e+02	&	$=$	&	1.55e+03$\pm$6.28e+02	&	$=$	&	7.63e+02$\pm$3.64e+02	&	$-$	&	3.07e+03$\pm$1.23e+03	&	$+$	\\ \hline
6	&	1.10e+07$\pm$1.24e+07	&	3.11e+07$\pm$1.07e+07	&	$+$	&	2.04e+07$\pm$1.04e+07	&	$+$	&	2.25e+07$\pm$1.51e+07	&	$+$	&	3.74e+07$\pm$1.93e+07	&	$+$	&	4.20e+07$\pm$7.09e+07	&	$=$	\\ \hline
7	&	5.82e+04$\pm$1.61e+05	&	2.25e+05$\pm$6.13e+04	&	$+$	&	1.56e+05$\pm$5.12e+04	&	$+$	&	4.53e+05$\pm$6.93e+04	&	$+$	&	6.70e+05$\pm$9.16e+04	&	$+$	&	4.53e+03$\pm$7.79e+03	&	$-$	\\ \hline
8	&	1.16e+09$\pm$3.41e+09	&	9.63e+09$\pm$2.55e+09	&	$+$	&	6.09e+09$\pm$3.89e+09	&	$+$	&	5.79e+10$\pm$1.69e+10	&	$+$	&	1.07e+11$\pm$1.54e+10	&	$+$	&	9.10e+06$\pm$2.55e+07	&	$-$	\\ \hline
9	&	2.85e+07$\pm$8.54e+07	&	7.35e+09$\pm$2.53e+09	&	$+$	&	3.66e+09$\pm$2.98e+09	&	$+$	&	3.95e+10$\pm$9.54e+09	&	$+$	&	1.16e+11$\pm$1.84e+10	&	$+$	&	2.86e+06$\pm$1.25e+07	&	$-$	\\ \hline
10	&	2.10e+08$\pm$5.82e+08	&	5.39e+08$\pm$1.61e+08	&	$+$	&	3.80e+08$\pm$1.47e+08	&	$=$	&	8.18e+08$\pm$2.53e+08	&	$+$	&	1.37e+09$\pm$2.94e+08	&	$+$	&	4.69e+07$\pm$5.60e+07	&	$-$	\\ \hline
11	&	1.51e+06$\pm$3.35e+06	&	3.30e+05$\pm$1.65e+05	&	$-$	&	3.17e+05$\pm$1.69e+05	&	$-$	&	6.38e+05$\pm$2.33e+05	&	$-$	&	2.74e+05$\pm$3.09e+04	&	$=$	&	1.52e+06$\pm$3.29e+06	&	$=$	\\ \hline
12	&	7.10e+13$\pm$2.64e+14	&	3.98e+14$\pm$1.02e+15	&	$=$	&	4.64e+13$\pm$7.43e+13	&	$=$	&	8.75e+14$\pm$1.84e+15	&	$+$	&	2.09e+17$\pm$4.10e+17	&	$+$	&	5.56e+16$\pm$1.71e+17	&	$+$	\\ \hline
13	&	1.15e+04$\pm$1.86e+04	&	3.36e+04$\pm$2.68e+03	&	$=$	&	2.61e+04$\pm$3.97e+03	&	$=$	&	5.30e+04$\pm$5.23e+03	&	$+$	&	6.48e+04$\pm$4.26e+03	&	$+$	&	7.33e+03$\pm$9.10e+03	&	$-$	\\ \hline
14	&	1.43e+04$\pm$3.37e+04	&	1.30e+04$\pm$3.81e+03	&	$=$	&	7.13e+03$\pm$3.90e+03	&	$-$	&	2.77e+04$\pm$1.05e+04	&	$=$	&	4.43e+04$\pm$6.97e+03	&	$+$	&	1.48e+02$\pm$1.43e+02	&	$-$	\\ \hline
15	&	1.66e+06$\pm$3.32e+06	&	1.50e+06$\pm$6.33e+05	&	$=$	&	7.63e+05$\pm$4.99e+05	&	$-$	&	3.96e+06$\pm$3.14e+06	&	$+$	&	2.28e+07$\pm$1.47e+07	&	$+$	&	7.64e+07$\pm$2.86e+08	&	$+$	\\ \hline
16	&	2.11e+02$\pm$1.13e+01	&	2.64e+03$\pm$4.55e+02	&	$+$	&	1.66e+03$\pm$5.55e+02	&	$+$	&	6.07e+03$\pm$1.03e+03	&	$+$	&	9.61e+03$\pm$1.16e+03	&	$+$	&	2.80e+02$\pm$3.16e+02	&	$=$	\\ \hline
17	&	1.73e+05$\pm$2.25e+05	&	5.14e+05$\pm$1.48e+05	&	$+$	&	3.85e+05$\pm$1.95e+05	&	$+$	&	1.88e+06$\pm$4.56e+05	&	$+$	&	3.27e+06$\pm$9.89e+05	&	$+$	&	1.14e+06$\pm$1.29e+06	&	$+$	\\ \hline
18	&	2.70e+05$\pm$3.24e+05	&	6.01e+05$\pm$2.31e+05	&	$+$	&	4.54e+05$\pm$3.55e+05	&	$=$	&	2.40e+06$\pm$8.46e+05	&	$+$	&	3.59e+06$\pm$9.43e+05	&	$+$	&	3.61e+05$\pm$8.52e+05	&	$=$	\\ \hline
19	&	6.85e+03$\pm$2.94e+04	&	1.96e+05$\pm$7.70e+04	&	$+$	&	8.20e+04$\pm$5.07e+04	&	$+$	&	8.91e+05$\pm$3.80e+05	&	$+$	&	2.51e+06$\pm$5.46e+05	&	$+$	&	4.34e+01$\pm$3.71e+00	&	$-$	\\ \hline
20	&	1.14e+07$\pm$2.83e+07	&	4.55e+07$\pm$9.60e+06	&	$+$	&	2.46e+07$\pm$9.70e+06	&	$=$	&	1.09e+08$\pm$2.14e+07	&	$+$	&	1.63e+08$\pm$1.86e+07	&	$+$	&	2.49e+05$\pm$7.58e+05	&	$-$	\\ \hline
21	&	4.51e+02$\pm$2.37e+02	&	2.38e+04$\pm$5.22e+03	&	$+$	&	1.38e+04$\pm$4.12e+03	&	$+$	&	6.61e+04$\pm$1.37e+04	&	$+$	&	9.72e+04$\pm$1.52e+04	&	$+$	&	3.96e+02$\pm$7.11e+00	&	$=$	\\ \hline
22	&	1.22e+02$\pm$2.12e+00	&	2.16e+04$\pm$3.28e+03	&	$+$	&	1.16e+04$\pm$3.31e+03	&	$+$	&	6.41e+04$\pm$1.28e+04	&	$+$	&	9.47e+04$\pm$1.37e+04	&	$+$	&	1.27e+02$\pm$5.11e-01	&	$=$	\\ \hline
23	&	1.90e+03$\pm$5.47e+03	&	2.32e+04$\pm$6.00e+03	&	$+$	&	1.46e+04$\pm$5.36e+03	&	$+$	&	6.51e+04$\pm$1.35e+04	&	$+$	&	1.01e+05$\pm$1.47e+04	&	$+$	&	2.21e+02$\pm$1.70e+01	&	$-$	\\ \hline
24	&	7.36e+06$\pm$3.21e+07	&	2.31e+08$\pm$5.53e+07	&	$+$	&	1.27e+08$\pm$3.78e+07	&	$+$	&	6.05e+08$\pm$8.12e+07	&	$+$	&	9.74e+08$\pm$9.47e+07	&	$+$	&	9.13e+06$\pm$3.94e+07	&	$+$	\\ \hline
\end{tabular}
\label{T:D100NP6}
\end{center}
\end{table*}
\begin{table*}[!htbp]
\captionsetup{font=footnotesize}
\tiny
\caption{Error and standard deviation of the proposed OEMDE algorithm versus state of the art. The \textit{W} represents the Wilcoxon statistical test results compared to OEMDE. The signs `$+$' and `$-$' show significance of the OEMDE over the competitive algorithm and vice-versa, respectively; `$=$' shows neither is significant. $\textbf{D=200}$ and $N_{P}=6$.}
\vspace{-0.3cm}
\begin{center}
\begin{tabular}{|c|c|c|c|c|c|c|c|c|c|c|c|}
\hline
\multirow{1}{*}{$F$} & \multicolumn{1}{c|}{OEMDE}  & \multicolumn{2}{c|}{EMDE} &  \multicolumn{2}{c|}{OIEMDE} & \multicolumn{2}{c|}{PAjDE} & \multicolumn{2}{c|}{$\mu$JADE}& \multicolumn{2}{c|}{ODE}\\ \hhline{~-----------}
 & \multicolumn{1}{c|}{Error}  & \multicolumn{1}{c|}{Error} &\multicolumn{1}{c|}{W} &  \multicolumn{1}{c|}{Error} &\multicolumn{1}{c|}{W} &\multicolumn{1}{c|}{Error} &\multicolumn{1}{c|}{W} & \multicolumn{1}{c|}{Error} &\multicolumn{1}{c|}{W}& \multicolumn{1}{c|}{Error} &\multicolumn{1}{c|}{W}\\
\hline
\hline
1	&	2.14e+03$\pm$4.52e+03	&	8.83e+04$\pm$1.02e+04		&	$+$	&	5.58e+04$\pm$1.97e+04	&	$+$	&	2.45e+05$\pm$3.56e+04	&	$+$	&	3.36e+05$\pm$2.62e+04	&	$+$	&	9.39e+02$\pm$2.90e+01	&	$-$	\\ \hline
2	&	6.95e+08$\pm$1.62e+09	&	3.73e+09$\pm$9.10e+08		&	$+$	&	2.50e+09$\pm$1.04e+09	&	$+$	&	6.38e+09$\pm$1.81e+09	&	$+$	&	8.92e+09$\pm$1.76e+09	&	$+$	&	5.74e+08$\pm$2.13e+09	&	$=$	\\ \hline
3	&	4.05e+07$\pm$8.58e+07	&	1.11e+08$\pm$7.75e+07		&	$+$	&	2.05e+08$\pm$2.89e+08	&	$+$	&	5.52e+09$\pm$4.49e+09	&	$+$	&	2.61e+09$\pm$4.43e+09	&	$+$	&	2.32e+09$\pm$6.08e+09	&	$+$	\\ \hline
4	&	1.85e+06$\pm$5.42e+06	&	1.23e+07$\pm$2.18e+06		&	$+$	&	9.50e+06$\pm$2.60e+06	&	$+$	&	3.36e+07$\pm$4.59e+06	&	$+$	&	4.62e+07$\pm$2.79e+06	&	$+$	&	2.76e+06$\pm$9.84e+06	&	$=$	\\ \hline
5	&	4.13e+03$\pm$1.30e+03	&	5.60e+03$\pm$8.84e+02		&	$+$	&	4.86e+03$\pm$5.89e+02	&	$+$	&	5.20e+03$\pm$1.40e+03	&	$+$	&	4.82e+03$\pm$9.33e+02	&	$=$	&	5.08e+03$\pm$2.96e+03	&	$=$	\\ \hline
6	&	3.32e+07$\pm$4.67e+07	&	1.28e+08$\pm$2.68e+07		&	$+$	&	9.17e+07$\pm$2.03e+07	&	$+$	&	1.59e+08$\pm$3.24e+07	&	$+$	&	2.63e+08$\pm$5.60e+07	&	$+$	&	6.22e+07$\pm$2.32e+08	&	$=$	\\ \hline
7	&	5.36e+04$\pm$1.24e+05	&	6.49e+05$\pm$1.17e+05		&	$+$	&	3.80e+05$\pm$7.66e+04	&	$+$	&	1.40e+06$\pm$1.88e+05	&	$+$	&	1.87e+06$\pm$1.78e+05	&	$+$	&	9.47e+03$\pm$3.29e+03	&	$-$	\\ \hline
8	&	1.33e+09$\pm$5.62e+09	&	1.15e+11$\pm$2.73e+10		&	$+$	&	4.79e+10$\pm$3.08e+10	&	$+$	&	8.02e+11$\pm$1.56e+11	&	$+$	&	1.29e+12$\pm$1.55e+11	&	$+$	&	1.77e+07$\pm$3.83e+07	&	$-$	\\ \hline
9	&	1.27e+08$\pm$4.31e+08	&	1.04e+11$\pm$2.67e+10		&	$+$	&	4.78e+10$\pm$2.51e+10	&	$+$	&	6.73e+11$\pm$1.87e+11	&	$+$	&	1.43e+12$\pm$2.33e+11	&	$+$	&	3.84e+06$\pm$1.67e+07	&	$-$	\\ \hline
10	&	4.90e+07$\pm$8.11e+07	&	1.88e+09$\pm$2.26e+08		&	$+$	&	1.15e+09$\pm$3.51e+08	&	$+$	&	3.26e+09$\pm$6.43e+08	&	$+$	&	4.79e+09$\pm$6.59e+08	&	$+$	&	7.98e+08$\pm$3.31e+09	&	$+$	\\ \hline
11	&	2.07e+06$\pm$4.05e+06	&	5.19e+05$\pm$2.45e+05		&	$-$	&	3.47e+05$\pm$1.96e+05	&	$-$	&	1.28e+06$\pm$2.76e+05	&	$=$	&	6.00e+05$\pm$4.81e+04	&	$=$	&	1.06e+06$\pm$1.37e+06	&	$=$	\\ \hline
12	&	1.02e+16$\pm$4.02e+16	&	2.35e+16$\pm$4.91e+16		&	$=$	&	2.50e+15$\pm$4.01e+15	&	$-$	&	9.64e+17$\pm$2.86e+18	&	$+$	&	2.77e+20$\pm$4.68e+20	&	$+$	&	6.01e+20$\pm$1.67e+21	&	$-$	\\ \hline
13	&	9.97e+03$\pm$1.27e+04	&	5.79e+04$\pm$3.87e+03		&	$+$	&	4.58e+04$\pm$6.86e+03	&	$+$	&	9.15e+04$\pm$5.68e+03	&	$+$	&	1.06e+05$\pm$3.93e+03	&	$+$	&	6.47e+03$\pm$2.87e+02	&	$=$	\\ \hline
14	&	1.83e+04$\pm$6.72e+04	&	3.65e+04$\pm$7.22e+03		&	$=$	&	1.69e+04$\pm$8.08e+03	&	$=$	&	9.63e+04$\pm$3.00e+04	&	$+$	&	1.37e+05$\pm$1.84e+04	&	$+$	&	1.30e+02$\pm$3.99e+01	&	$-$	\\ \hline
15	&	2.30e+06$\pm$5.15e+06	&	9.71e+06$\pm$8.18e+06		&	$+$	&	5.02e+06$\pm$3.19e+06	&	$=$	&	4.93e+07$\pm$2.63e+07	&	$+$	&	5.45e+08$\pm$4.06e+08	&	$+$	&	7.06e+07$\pm$1.92e+08	&	$+$	\\ \hline
16	&	2.79e+02$\pm$4.59e+02	&	3.37e+03$\pm$3.99e+02		&	$+$	&	1.77e+03$\pm$7.64e+02	&	$+$	&	9.14e+03$\pm$1.09e+03	&	$+$	&	1.35e+04$\pm$7.57e+02	&	$+$	&	2.02e+02$\pm$9.25e+00	&	$=$	\\ \hline
17	&	6.01e+05$\pm$6.04e+05	&	1.31e+06$\pm$3.00e+05		&	$+$	&	1.07e+06$\pm$3.25e+05	&	$+$	&	5.55e+06$\pm$9.97e+05	&	$+$	&	1.06e+07$\pm$4.99e+06	&	$+$	&	3.93e+04$\pm$1.71e+05	&	$-$	\\ \hline
18	&	5.75e+05$\pm$8.37e+05	&	1.88e+06$\pm$8.97e+05		&	$+$	&	1.29e+06$\pm$5.73e+05	&	$+$	&	6.63e+06$\pm$2.50e+06	&	$+$	&	2.34e+07$\pm$1.16e+07	&	$+$	&	5.42e+06$\pm$1.89e+07	&	$+$	\\ \hline
19	&	4.78e+01$\pm$7.93e-01	&	1.29e+06$\pm$2.99e+05		&	$+$	&	6.48e+05$\pm$4.02e+05	&	$+$	&	7.93e+06$\pm$2.53e+06	&	$+$	&	1.73e+07$\pm$1.52e+06	&	$+$	&	1.13e+02$\pm$3.05e+02	&	$+$	\\ \hline
20	&	7.51e+06$\pm$1.70e+07	&	1.25e+08$\pm$1.53e+07		&	$+$	&	8.46e+07$\pm$2.64e+07	&	$+$	&	3.21e+08$\pm$4.37e+07	&	$+$	&	4.58e+08$\pm$4.37e+07	&	$+$	&	2.51e+07$\pm$1.07e+08	&	$+$	\\ \hline
21	&	4.22e+02$\pm$1.13e+02	&	6.08e+04$\pm$6.89e+03		&	$+$	&	3.87e+04$\pm$1.23e+04	&	$+$	&	1.86e+05$\pm$1.86e+04	&	$+$	&	2.78e+05$\pm$2.39e+04	&	$+$	&	3.99e+02$\pm$2.44e+01	&	$=$	\\ \hline
22	&	3.48e+02$\pm$9.32e+02	&	6.12e+04$\pm$5.74e+03		&	$+$	&	4.03e+04$\pm$1.13e+04	&	$+$	&	1.86e+05$\pm$2.37e+04	&	$+$	&	2.80e+05$\pm$1.81e+04	&	$+$	&	1.66e+02$\pm$1.69e+02	&	$=$	\\ \hline
23	&	9.87e+02$\pm$3.32e+03	&	6.15e+04$\pm$7.60e+03		&	$+$	&	3.96e+04$\pm$1.22e+04	&	$+$	&	1.79e+05$\pm$1.90e+04	&	$+$	&	2.81e+05$\pm$1.72e+04	&	$+$	&	2.15e+02$\pm$6.86e-01	&	$=$	\\ \hline
24	&	1.62e+06$\pm$5.05e+06	&	5.96e+08$\pm$8.59e+07		&	$+$	&	3.18e+08$\pm$8.75e+07	&	$+$	&	1.84e+09$\pm$1.98e+08	&	$+$	&	2.79e+09$\pm$2.08e+08	&	$+$	&	3.02e+03$\pm$1.70e+02	&	$-$	\\ \hline
\end{tabular}
\label{T:D200NP6}
\end{center}
\end{table*}
\begin{table*}[!htp]
\captionsetup{font=footnotesize}
\tiny
\caption{Error and standard deviation of the proposed OEMDE algorithm versus state of the art.The \textit{W} represents the Wilcoxon statistical test results compared to OEMDE. The signs `$+$' and `$-$' show significance of the OEMDE over the competitive algorithm and vice-versa, respectively; `$=$' shows neither is significant. $\textbf{D=500}$ and $N_{P}=6$.}
\vspace{-0.3cm}
\begin{center}
\begin{tabular}{|c|c|c|c|c|c|c|c|c|c|c|c|}
\hline
\multirow{1}{*}{$F$} & \multicolumn{1}{c|}{OEMDE}  & \multicolumn{2}{c|}{EMDE} &  \multicolumn{2}{c|}{OIEMDE} & \multicolumn{2}{c|}{PAjDE} & \multicolumn{2}{c|}{$\mu$JADE}& \multicolumn{2}{c|}{ODE}\\ \hhline{~-----------}
 & \multicolumn{1}{c|}{Error}  & \multicolumn{1}{c|}{Error} &\multicolumn{1}{c|}{W} &  \multicolumn{1}{c|}{Error} &\multicolumn{1}{c|}{W} &\multicolumn{1}{c|}{Error} &\multicolumn{1}{c|}{W} & \multicolumn{1}{c|}{Error} &\multicolumn{1}{c|}{W}& \multicolumn{1}{c|}{Error} &\multicolumn{1}{c|}{W}\\
\hline
\hline
																							
1	&	3.00e+03$\pm$2.05e+03	&	2.78e+05$\pm$2.48e+04	&	$+$	&	1.79e+05$\pm$4.56e+04	&	$+$	&	8.43e+05$\pm$6.78e+04	&	$+$	&	1.00e+06$\pm$3.61e+04	&	$+$	&	2.67e+03$\pm$1.04e+03	&	$=$	\\ \hline
2	&	1.37e+09$\pm$4.84e+09	&	1.45e+10$\pm$2.05e+09	&	$+$	&	7.97e+09$\pm$2.13e+09	&	$+$	&	3.46e+10$\pm$4.15e+09	&	$+$	&	4.14e+10$\pm$3.93e+09	&	$+$	&	1.67e+08$\pm$9.53e+06	&	$-$	\\ \hline
3	&	4.13e+08$\pm$1.04e+09	&	1.15e+09$\pm$9.36e+08	&	$+$	&	5.17e+08$\pm$4.22e+08	&	$=$	&	1.13e+11$\pm$9.00e+10	&	$+$	&	5.99e+10$\pm$3.00e+10	&	$+$	&	3.96e+11$\pm$9.21e+11	&	$+$	\\ \hline
4	&	4.75e+05$\pm$8.22e+05	&	4.15e+07$\pm$5.07e+06	&	$+$	&	3.19e+07$\pm$7.90e+06	&	$+$	&	1.25e+08$\pm$1.41e+07	&	$+$	&	1.51e+08$\pm$1.37e+07	&	$+$	&	2.57e+07$\pm$7.68e+07	&	$+$	\\ \hline
5	&	1.13e+04$\pm$4.37e+03	&	1.93e+04$\pm$9.43e+02	&	$+$	&	1.79e+04$\pm$1.43e+03	&	$+$	&	2.32e+04$\pm$2.00e+03	&	$+$	&	2.59e+04$\pm$1.74e+03	&	$+$	&	1.08e+04$\pm$5.32e+03	&	$=$	\\ \hline
6	&	7.74e+07$\pm$1.58e+08	&	4.16e+08$\pm$5.45e+07	&	$+$	&	3.49e+08$\pm$5.07e+07	&	$+$	&	8.72e+08$\pm$8.62e+07	&	$+$	&	1.17e+09$\pm$1.39e+08	&	$+$	&	1.47e+08$\pm$6.00e+08	&	$+$	\\ \hline
7	&	2.40e+04$\pm$5.46e+03	&	2.17e+06$\pm$2.76e+05	&	$+$	&	1.43e+06$\pm$2.62e+05	&	$+$	&	5.26e+06$\pm$5.40e+05	&	$+$	&	6.62e+06$\pm$4.92e+05	&	$+$	&	2.23e+04$\pm$4.62e+03	&	$=$	\\ \hline
8	&	2.05e+11$\pm$8.87e+11	&	2.50e+12$\pm$4.92e+11	&	$+$	&	1.34e+12$\pm$5.42e+11	&	$+$	&	2.08e+13$\pm$1.73e+12	&	$+$	&	2.66e+13$\pm$1.63e+12	&	$+$	&	9.03e+07$\pm$1.16e+08	&	$-$	\\ \hline
9	&	2.45e+11$\pm$1.07e+12	&	2.47e+12$\pm$4.78e+11	&	$+$	&	1.05e+12$\pm$4.62e+11	&	$+$	&	2.02e+13$\pm$2.57e+12	&	$+$	&	3.16e+13$\pm$3.18e+12	&	$+$	&	6.24e+04$\pm$2.56e+05	&	$-$	\\ \hline
10	&	2.13e+09$\pm$6.30e+09	&	9.57e+09$\pm$1.92e+09	&	$+$	&	6.53e+09$\pm$1.52e+09	&	$+$	&	1.96e+10$\pm$3.50e+09	&	$+$	&	2.38e+10$\pm$1.93e+09	&	$+$	&	1.74e+08$\pm$1.03e+07	&	$-$	\\ \hline
11	&	1.88e+06$\pm$4.11e+06	&	1.09e+06$\pm$6.59e+05	&	$=$	&	1.07e+06$\pm$5.83e+05	&	$=$	&	2.32e+06$\pm$4.78e+05	&	$=$	&	1.45e+06$\pm$1.25e+05	&	$=$	&	1.12e+06$\pm$1.53e+06	&	$=$	\\ \hline
12	&	8.45e+17$\pm$2.01e+18	&	2.62e+18$\pm$3.43e+18	&	$+$	&	5.76e+17$\pm$1.13e+18	&	$=$	&	5.53e+21$\pm$6.38e+21	&	$+$	&	1.21e+24$\pm$2.97e+24	&	$+$	&	2.37e+13$\pm$1.03e+14	&	$-$	\\ \hline
13	&	1.32e+04$\pm$1.19e+04	&	9.99e+04$\pm$5.15e+03	&	$+$	&	8.64e+04$\pm$1.08e+04	&	$+$	&	1.75e+05$\pm$9.05e+03	&	$+$	&	1.90e+05$\pm$5.61e+03	&	$+$	&	1.03e+04$\pm$2.11e+02	&	$=$	\\ \hline
14	&	3.62e+04$\pm$1.08e+05	&	1.11e+05$\pm$1.75e+04	&	$+$	&	5.92e+04$\pm$2.14e+04	&	$=$	&	3.13e+05$\pm$6.32e+04	&	$+$	&	4.44e+05$\pm$4.33e+04	&	$+$	&	5.86e+03$\pm$2.42e+04	&	$-$	\\ \hline
15	&	1.97e+07$\pm$5.20e+07	&	1.30e+08$\pm$7.17e+07	&	$+$	&	5.54e+07$\pm$4.19e+07	&	$+$	&	1.87e+09$\pm$1.25e+09	&	$+$	&	2.18e+10$\pm$2.12e+10	&	$+$	&	3.27e+04$\pm$5.40e+04	&	$-$	\\ \hline
16	&	6.97e+02$\pm$2.08e+03	&	4.01e+03$\pm$4.44e+02	&	$+$	&	2.82e+03$\pm$6.66e+02	&	$+$	&	1.32e+04$\pm$9.29e+02	&	$+$	&	1.59e+04$\pm$5.44e+02	&	$+$	&	1.99e+02$\pm$5.62e+00	&	$=$	\\ \hline
17	&	9.43e+05$\pm$1.74e+06	&	4.04e+06$\pm$6.34e+05	&	$+$	&	2.78e+06$\pm$7.74e+05	&	$+$	&	1.99e+07$\pm$2.73e+06	&	$+$	&	2.20e+08$\pm$2.19e+08	&	$+$	&	1.15e+07$\pm$3.48e+07	&	$+$	\\ \hline
18	&	9.22e+06$\pm$2.85e+07	&	5.88e+06$\pm$2.02e+06	&	$=$	&	3.86e+06$\pm$1.91e+06	&	$=$	&	5.72e+07$\pm$2.79e+07	&	$+$	&	7.90e+08$\pm$4.19e+08	&	$+$	&	4.46e+07$\pm$1.75e+08	&	$+$	\\ \hline
19	&	1.41e+04$\pm$5.44e+04	&	1.29e+07$\pm$2.38e+06	&	$+$	&	5.23e+06$\pm$2.15e+06	&	$+$	&	1.03e+08$\pm$1.50e+07	&	$+$	&	1.57e+08$\pm$1.98e+07	&	$+$	&	1.28e+03$\pm$5.02e+03	&	$-$	\\ \hline
20	&	5.10e+07$\pm$1.43e+08	&	3.90e+08$\pm$3.64e+07	&	$+$	&	2.42e+08$\pm$6.46e+07	&	$+$	&	1.20e+09$\pm$8.96e+07	&	$+$	&	1.48e+09$\pm$7.82e+07	&	$+$	&	4.72e+05$\pm$1.76e+05	&	$-$	\\ \hline
21	&	3.95e+02$\pm$1.20e+01	&	1.91e+05$\pm$1.61e+04	&	$+$	&	1.30e+05$\pm$3.83e+04	&	$+$	&	6.80e+05$\pm$5.04e+04	&	$+$	&	8.16e+05$\pm$4.01e+04	&	$+$	&	3.94e+02$\pm$2.35e-01	&	$=$	\\ \hline
22	&	1.25e+02$\pm$1.16e+00	&	1.95e+05$\pm$1.94e+04	&	$+$	&	1.28e+05$\pm$4.04e+04	&	$+$	&	6.64e+05$\pm$4.79e+04	&	$+$	&	8.26e+05$\pm$4.32e+04	&	$+$	&	1.26e+02$\pm$1.07e-01	&	$+$	\\ \hline
23	&	1.23e+32$\pm$2.48e+31	&	8.99e+31$\pm$7.47e+30	&	$-$	&	1.10e+32$\pm$3.63e+31	&	$=$	&	5.40e+32$\pm$3.33e+31	&	$=$	&	8.40e+32$\pm$2.67e+31	&	$=$	&	1.78e+32$\pm$5.03e+31	&	$+$	\\ \hline
24	&	7.78e+06$\pm$3.11e+07	&	1.94e+09$\pm$2.20e+08	&	$+$	&	1.19e+09$\pm$2.81e+08	&	$+$	&	6.47e+09$\pm$4.48e+08	&	$+$	&	8.11e+09$\pm$4.25e+08	&	$+$	&	7.77e+03$\pm$2.36e+02	&	$-$	\\ \hline

\end{tabular}
\label{T:D500NP6}
\end{center}
\end{table*}
\begin{table*}[!htp]
\captionsetup{font=footnotesize}
\tiny
\caption{Error and standard deviation of the proposed OEMDE algorithm versus state of the art. The \textit{W} represents the Wilcoxon statistical test results compared to OEMDE. The signs `$+$' and `$-$' show significance of the OEMDE over the competitive algorithm and vice-versa, respectively; `$=$' shows neither is significant. $\textbf{D=1000}$ and $N_{P}=6$.}
\vspace{-0.3cm}
\begin{center}
\begin{tabular}{|c|c|c|c|c|c|c|c|c|c|c|c|}
\hline
\multirow{1}{*}{$F$} & \multicolumn{1}{c|}{OEMDE}  & \multicolumn{2}{c|}{EMDE} &  \multicolumn{2}{c|}{OIEMDE} & \multicolumn{2}{c|}{PAjDE} & \multicolumn{2}{c|}{$\mu$JADE}& \multicolumn{2}{c|}{ODE}\\ \hhline{~-----------}
 & \multicolumn{1}{c|}{Error}  & \multicolumn{1}{c|}{Error} &\multicolumn{1}{c|}{W} &  \multicolumn{1}{c|}{Error} &\multicolumn{1}{c|}{W} &\multicolumn{1}{c|}{Error} &\multicolumn{1}{c|}{W} & \multicolumn{1}{c|}{Error} &\multicolumn{1}{c|}{W}& \multicolumn{1}{c|}{Error} &\multicolumn{1}{c|}{W}\\
\hline
\hline
1	&	5.77e+04$\pm$1.15e+05	&	6.09e+05$\pm$5.09e+04	&	$+$	&	4.05e+05$\pm$7.57e+04	&	$+$	&	1.88e+06$\pm$1.76e+05	&	$+$	&	2.08e+06$\pm$5.80e+04	&	$+$	&	9.75e+03$\pm$2.12e+04	&	$-$		\\ \hline
2	&	2.09e+09$\pm$6.37e+09	&	3.40e+10$\pm$3.38e+09	&	$+$	&	2.40e+10$\pm$6.66e+09	&	$+$	&	8.49e+10$\pm$9.38e+09	&	$+$	&	1.16e+11$\pm$1.03e+10	&	$+$	&	1.21e+09$\pm$3.61e+09	&	$=$		\\ \hline
3	&	1.75e+09$\pm$5.55e+09	&	3.50e+09$\pm$2.88e+09	&	$=$	&	1.44e+09$\pm$1.05e+09	&	$=$	&	4.05e+11$\pm$1.64e+11	&	$+$	&	5.48e+11$\pm$1.92e+11	&	$+$	&	5.72e+05$\pm$1.66e+06	&	$+$		\\ \hline
4	&	1.44e+07$\pm$3.78e+07	&	9.06e+07$\pm$7.43e+06	&	$+$	&	6.95e+07$\pm$1.78e+07	&	$+$	&	2.95e+08$\pm$1.69e+07	&	$+$	&	3.46e+08$\pm$1.94e+07	&	$+$	&	4.11e+07$\pm$1.29e+08	&	$=$		\\ \hline
5	&	1.93e+04$\pm$5.50e+03	&	4.28e+04$\pm$1.95e+03	&	$+$	&	4.05e+04$\pm$2.90e+03	&	$+$	&	5.98e+04$\pm$3.16e+03	&	$+$	&	6.87e+04$\pm$4.16e+03	&	$+$	&	2.52e+04$\pm$1.39e+04	&	$=$		\\ \hline
6	&	3.30e+07$\pm$2.40e+07	&	9.67e+08$\pm$8.63e+07	&	$+$	&	6.62e+08$\pm$1.26e+08	&	$+$	&	2.66e+09$\pm$2.78e+08	&	$+$	&	2.90e+09$\pm$1.37e+07	&	$+$	&	8.28e+08$\pm$1.81e+09	&	$+$		\\ \hline
7	&	6.07e+04$\pm$1.51e+04	&	4.84e+06$\pm$4.41e+05	&	$+$	&	3.99e+06$\pm$1.22e+06	&	$+$	&	1.47e+07$\pm$1.00e+06	&	$+$	&	1.69e+07$\pm$5.80e+05	&	$+$	&	7.82e+04$\pm$6.59e+04	&	$=$		\\ \hline
8	&	1.04e+11$\pm$6.05e+11	&	2.22e+13$\pm$3.16e+12	&	$+$	&	9.71e+12$\pm$3.96e+12	&	$+$	&	2.14e+14$\pm$2.11e+13	&	$+$	&	2.28e+14$\pm$1.59e+13	&	$+$	&	5.13e+08$\pm$2.06e+08	&	$+$		\\ \hline
9	&	3.50e+09$\pm$9.84e+09	&	2.45e+13$\pm$3.25e+12	&	$+$	&	1.29e+13$\pm$4.73e+12	&	$+$	&	2.40e+14$\pm$3.37e+13	&	$+$	&	2.78e+14$\pm$1.57e+13	&	$+$	&	5.91e+13$\pm$1.96e+14	&	$+$		\\ \hline
10	&	3.07e+08$\pm$1.20e+08	&	2.62e+10$\pm$3.25e+09	&	$+$	&	1.57e+10$\pm$3.70e+09	&	$+$	&	5.89e+10$\pm$8.86e+09	&	$+$	&	7.28e+10$\pm$4.33e+09	&	$+$	&	1.96e+10$\pm$6.02e+10	&	$+$		\\ \hline
11	&	2.06e+06$\pm$1.57e+06	&	2.61e+06$\pm$1.12e+06	&	$=$	&	1.89e+06$\pm$1.18e+06	&	$=$	&	4.46e+06$\pm$9.30e+05	&	$+$	&	2.84e+06$\pm$2.21e+05	&	$=$	&	3.58e+06$\pm$5.89e+06	&	$=$		\\ \hline
12	&	2.42e+10$\pm$2.51e+10	&	1.74e+20$\pm$4.48e+20	&	$=$	&	3.04e+19$\pm$7.94e+19	&	$=$	&	5.32e+23$\pm$5.91e+23	&	$+$	&	3.16e+26$\pm$3.56e+26	&	$+$	&	1.58e+14$\pm$5.24e+14	&	$+$		\\ \hline
13	&	1.44e+04$\pm$2.30e+02	&	1.52e+05$\pm$6.94e+03	&	$+$	&	1.27e+05$\pm$1.15e+04	&	$+$	&	2.64e+05$\pm$4.38e+03	&	$+$	&	2.86e+05$\pm$1.50e+03	&	$+$	&	1.49e+04$\pm$4.28e+01	&	$+$		\\ \hline
14	&	2.45e+04$\pm$1.02e+05	&	2.04e+05$\pm$2.50e+04	&	$+$	&	1.46e+05$\pm$6.31e+04	&	$+$	&	7.24e+05$\pm$6.29e+04	&	$+$	&	1.06e+06$\pm$1.16e+05	&	$+$	&	4.50e+02$\pm$1.29e+02	&	$-$		\\ \hline
15	&	1.26e+07$\pm$2.17e+07	&	8.11e+08$\pm$3.99e+08	&	$+$	&	3.68e+08$\pm$1.72e+08	&	$+$	&	6.02e+10$\pm$4.01e+10	&	$+$	&	1.95e+11$\pm$1.52e+11	&	$=$	&	4.71e+04$\pm$2.20e+04	&	$-$		\\ \hline
16	&	1.99e+02$\pm$3.45e+00	&	4.25e+03$\pm$4.51e+02	&	$+$	&	3.04e+03$\pm$7.43e+01	&	$+$	&	1.57e+04$\pm$1.54e+03	&	$+$	&	1.73e+04$\pm$1.10e+03	&	$+$	&	5.83e+02$\pm$8.65e+02	&	$=$		\\ \hline
17	&	2.49e+06$\pm$3.53e+06	&	8.95e+06$\pm$2.16e+06	&	$+$	&	6.24e+06$\pm$1.86e+06	&	$+$	&	1.59e+08$\pm$5.73e+07	&	$+$	&	1.59e+09$\pm$8.57e+08	&	$+$	&	1.64e+01$\pm$1.34e+00	&	$-$		\\ \hline
18	&	3.83e+06$\pm$5.09e+06	&	1.95e+07$\pm$1.06e+07	&	$+$	&	1.00e+07$\pm$2.01e+06	&	$=$	&	2.91e+08$\pm$6.02e+07	&	$+$	&	3.70e+09$\pm$1.18e+09	&	$+$	&	4.98e+01$\pm$5.45e+00	&	$-$		\\ \hline
19	&	3.37e+07$\pm$8.25e+07	&	5.81e+07$\pm$7.05e+06	&	$=$	&	3.08e+07$\pm$1.11e+07	&	$=$	&	5.77e+08$\pm$5.71e+07	&	$+$	&	7.59e+08$\pm$5.60e+07	&	$+$	&	4.35e+01$\pm$3.95e+00	&	$-$		\\ \hline
20	&	1.27e+08$\pm$3.85e+08	&	8.50e+08$\pm$7.09e+07	&	$+$	&	7.12e+08$\pm$1.39e+08	&	$+$	&	2.76e+09$\pm$1.67e+08	&	$+$	&	3.23e+09$\pm$1.32e+08	&	$+$	&	8.43e+05$\pm$1.28e+05	&	$-$		\\ \hline
21	&	3.94e+02$\pm$2.17e-01	&	4.00e+05$\pm$4.41e+04	&	$+$	&	2.72e+05$\pm$6.62e+04	&	$+$	&	1.53e+06$\pm$1.27e+05	&	$+$	&	1.66e+06$\pm$6.69e+04	&	$+$	&	3.99e+02$\pm$9.17e+00	&	$=$		\\ \hline
22	&	1.26e+02$\pm$1.24e-01	&	4.08e+05$\pm$3.74e+04	&	$+$	&	3.26e+05$\pm$7.25e+04	&	$+$	&	1.58e+06$\pm$1.25e+05	&	$+$	&	1.74e+06$\pm$5.30e+04	&	$+$	&	1.27e+02$\pm$7.42e-02	&	$=$		\\ \hline
23	&	1.36e+33$\pm$5.50e+32	&	8.51e+32$\pm$2.49e+32	&	$-$	&	9.55e+32$\pm$3.83e+32	&	$=$	&	7.53e+33$\pm$3.60e+32	&	$=$	&	4.33e+33$\pm$4.70e+32	&	$=$	&	1.74e+33$\pm$1.11e+32	&	$=$		\\ \hline
24	&	1.56e+04$\pm$1.58e+02	&	4.21e+09$\pm$4.13e+08	&	$+$	&	2.38e+09$\pm$2.34e+08	&	$+$	&	1.49e+10$\pm$8.46e+08	&	$+$	&	1.72e+10$\pm$2.72e+08	&	$+$	&	1.58e+04$\pm$3.82e+02	&	$=$		\\ \hline

\end{tabular}
\label{T:D1000NP6}
\end{center}
\end{table*}

\subsection{Performance Analysis}
The performance results of the OEMDE versus EMDE, OIEMDE, PAjDE, $\mu$JADE, and ODE with $N_{P}=6$ for $D\in\{30,50,100,200,500,1000\}$ are presented in Table~\ref{T:summary}. The Wilcoxon statistical test is performed to find significant results. Due to space limitations, the error and standard deviation values per objective function for all dimensions except 30 are presented in Tables~\ref{T:D100NP6}-\ref{T:D1000NP6}.

Results in Table~\ref{T:summary} show that as the problem dimensionality increases OEMDE achieves more success rates compared to EMDE, OIEMDE, and PAjDE. We can observe similar success versus $\mu$JADE and ODE, but with more neutral rates, particularly for $D=1000$. For this dimension, OEMDE and ODE have similar performance, with 7 neutral significance and 7 positive significance for ODE and OEMDE. This is while OEMDE has much better performance comparing to PAjDE, with 22 positive significance. Error value of algorithms in Table~\ref{T:summary} is plotted versus number of function calls in Figures~\ref{fig:d30} and \ref{fig:d1000} for $D\in\{30,1000\}$ and some objective functions $f_{1},f_{6},f_{15},f_{21}$ (due to lack of space). The plots clearly show that the OEMDE converges to a better candidate solution much faster than the other methods. For example for $D=30$ and $f_{1}$ the $\mu$JADE and PAjDE algorithms are trying to decrease the error value while the ODE and OEMDE already have arrived to a much lower error value. As problem dimensionality increases, similar convergence pattern is observable. For example, Figure~\ref{fig:d1000} for $f_{6}$, $f_{15}$,  and $f_{21}$, shows gradual exploration and exploitation of PAjDE and $\mu$JADE algorithms. This is while the increased diversity in OEMDE helps a faster convergence.

\section{Conclusion}
In this paper, an opposition-based ensemble version of micro-differential evolution (OEMDE) algorithm using a pool of mutation schemes is proposed. The OEMDE uses randomized mutation scale factor, such that a different mutation scale factor is assigned to each dimension of each individual. The OEMDE utilizes different mutation schemes for each individual of population, rather than a fixed mutation scheme. It enhances the diversity of population by giving a chance to opposite vectors of the population and involving them during the evolution. The objective is to increase the exploration ability of micro-DE algorithm and to decrease the chance of pre-mature convergence and stagnation by providing more diversity in the generated mutant vectors. The OEMDE is simple to implement and its performance is superior to the classical micro-DE algorithm. Its performance is on a par with ODE. However, OEMDE uses a smaller population size and does not require the setting of mutation scale factor and mutation scheme, which requires less trial and error for parameter setting. 

Diversity enhancement is an important factor in micro-DE. Other possibilities to increase diversity needs to be studied at different levels of algorithm for different population sizes and problem dimensionalities. The proposed work can also be extended to multi-objective and many-objective problems.

\bibliographystyle{ieeetr}
\bibliography{references} 
\end{document}